\documentclass[11pt]{amsart}
\usepackage{amsmath}
\usepackage{amssymb}
\usepackage{graphicx} 
\usepackage{subfigure}
\usepackage[margin=2cm]{geometry}
\usepackage{float}
\usepackage{url}
\usepackage[sort, numbers]{natbib}
\usepackage{hyperref}

\newcommand{\dsZ}{\mathbb{Z}}
\newcommand{\dsR}{\mathbb{R}}
\numberwithin{equation}{section}

\title{Wavelets Are All You Need for Autoregressive Image Generation}
\author{Wael Mattar, Idan Levy, Nir Sharon and Shai Dekel}

\begin{document}

\begin{abstract}
In this paper, we take a new approach to autoregressive image generation that is based on two main ingredients. The first is wavelet image coding, which allows to tokenize the visual details of an image from coarse to fine details by ordering the information starting with the most significant bits of the most significant wavelet coefficients. The second is a variant of a language transformer whose architecture is re-designed and optimized for token sequences in this `wavelet language'. The transformer learns the significant statistical correlations within a token sequence, which are the manifestations of well-known correlations between the wavelet subbands at various resolutions. We show experimental results with conditioning on the generation process. 
\end{abstract}

\maketitle

\section{Introduction}

The generation of high-resolution visual information is certainly one of the most remarkable achievements of modern-age artificial intelligence. One of the prominent methods is diffusion-based models \cite{dhariwal2021diffusion,saharia2022photorealistic, NEURIPS2020_4c5bcfec, rombach2022high,ramesh2022hierarchical}. In essence, diffusion models attempt to learn inversions of ill-posed operators, such as additive Gaussian noise, blurring, etc., so an image may be generated from random noisy or blurry seeds. 

Another line of research is designing autoregressive models, that apply the architecture of powerful Large Language Models (LLMs) \cite{vaswani2017attention, FormalTrans, tay2022efficient}. These autoregressive methods  \cite{esser2021taming,ramesh2021zero} convert the image pixel representation to a series of visual tokens and then apply generative language techniques. 

In this paper, we refine this line of research and provide a mathematically robust approach to the autoregressive image generation process. To this end, we reach out to a classic technique in image processing, specifically, wavelet image coding \cite{EZW,SPIHT,JPEG2000Book}. Wavelets \cite{Daubechies, DeVore, Mallat} are one of the main tools of modern approximation theory for nonlinear, adaptive approximation. The various wavelet transforms provide the means to transform an image into a representation that captures the essence of the visual information in a sparse way. Typically, the significant wavelet coefficients are a small fraction of the coefficients and represent important edge and texture information, while the insignificant coefficients with small absolute values are associated with smooth regions of the image. The goal of wavelet image compression methods, such as JPEG2000, is then to efficiently store the information of only the significant coefficients. In fact, the underlying method of the popular JPEG image compression algorithm \cite{Wallace},  invented in the 80s,  contains many elements of wavelet coding, where a local Discrete Cosine Transform, a precursor of wavelets, is used. However, in this paper, we leverage the progressive wavelet compression technique, a more advanced form of image compression. It creates a bit-stream where every bit corresponds to the next most important piece of visual information. Since we are generating images rather than decoding them from a compressed file, there is no need to create actual binary bit-streams, and using a `wavelet language' of a limited number of tokens is sufficient. 

Thus, our new approach to autoregressive image generation is based on two main ingredients. The first is progressive wavelet image coding, which allows to tokenize the visual information of an image from coarse to fine details. This can achieved using as few as 6 tokens, by ordering the information starting with the most significant bits of the most significant wavelet coefficients. The second ingredient is a variant of an NLP decoder-only transformer \cite{vaswani2017attention, FormalTrans, tay2022efficient} whose architecture was re-designed and optimized for token sequences in this `wavelet language' The transformer learns the significant statistical correlations within a token sequence, which are the manifestations of well-known correlations between the wavelet subbands at various resolutions \cite{mihcak1999spatially,803428,buccigrossi1999image}. During inference, this allows the generation of visually meaningful images from an initial random seed generated from sampled from the distribution of the scaling function coefficients at the lowest resolution.  

Using the wavelet autoregressive approach, where the `wavelet language' contains only a few tokens, provides many attractive features. The length of the token sequences during training or inference can be flexible, where longer sequences imply more detailed or higher-resolution images. Guiding the generative process using a class affiliation or text prompting is easily achieved by concatenating the corresponding vector representations to the tokens' vector representation of low dimension. Stochastic control using simple transformer inference techniques, allows to create from one textual prompt a diversity of corresponding images. Furthermore, since each token is associated with the local support in the image domain of the corresponding wavelet, one can switch the guidance during the generative process to allow different prompting for different regions.     

Our paper is organized as follows. We begin with a review of related work in Section \ref{sec:related}. In Section \ref{sec:waveletcoding}, we review wavelet image coding and explain how one may extract from the classical theory the ability to tokenize the visual information of images. In Section \ref{sec:GWT}, we focus on components of language transforms that we redesign to serve our special wavelet language. We further provide several methods that can be used to direct the generation process under certain conditions: class label and/or textual prompt. In Section \ref{sec:experiment}, we provide experimental results. Finally, in Section \ref{sec:discuss}, we discuss possible future applications of our method, such as multi-modality generation and compositions of blobs.

\section{Related work} \label{sec:related}

In this section, we first review the current state of the art in image generation. We then review some methods that apply wavelets as a frequency decomposition backbone for various aspects of style transfer, acceleration, and optimization of existing image generation methods, etc. 

Currently, many commercial solutions apply diffusion-based models \cite{dhariwal2021diffusion,saharia2022photorealistic, NEURIPS2020_4c5bcfec, rombach2022high,ramesh2022hierarchical,pmlr-v162-nichol22a}. In essence, diffusion models learn inversions of ill-posed operators, such as additive Gaussian noise, blurring, etc., so images may be generated from random noisy or blurry seeds. One then enforces various conditions on images created through the time steps of the inversion process so that the final generated image may correspond to a given text prompt. 

Recently, there is renewed interest in autoregressive methods with the hope that they will outperform the diffusion models. The methods of VQGAN \cite{esser2021taming} and DALL-E \cite{ramesh2021zero} along with \cite{wang2023images,lee2022autoregressive} utilize a visual tokenizer to discretize images into grids of 2D tokens, which are then flattened to a 1D sequence for autoregressive learning, mirroring the process of sequential language modeling. For example, in \cite{ramesh2021zero} a discrete variational autoencoder is trained to compress each $256 \times 256$ RGB image into a $32 \times 32$ grid of image tokens, where each such token can assume 8192 possible values. This creates a relatively short context sequence of $1024$ tokens, but with a vocabulary of 8192 word tokens. The TiTok method, recently introduced in \cite{32tokens}, shows how to combine Vision Transformers with the Vector-Quantization method to arrive at an autoregressive method that may use only 32 tokens. In comparison, our method may use only 6-7 tokens for any image resolution and any level of fine detail generation.

In contrast to typical raster-scan methods, where single tokens are sequentially predicted, the method of \cite{tian2024visual} provides an 
autoregressive learning algorithm based on predicting the image's next-scale, or next-resolution.

Some methods, such as \cite{zhu2023wdig,yu2021wavefill}, use wavelets as means for frequency decomposition representations for image inpainting,  style transfer, and generative adversarial network methods. Some works propose to use wavelets as part of diffusion methods \cite{waveletdiffusion1,guth2022wavelet} to speed up the diffusion approach by applying the denoising process in the wavelet regime. 

To the best of our knowledge, this is the first time wavelets are being used as the basis for autoregressive image generation.

\section{Elements of Wavelet Image Coding} \label{sec:waveletcoding}

In this section, we review some elements of wavelet image coding \cite{EZW,SPIHT,JPEG2000Book} that we use for our generative method.  Essentially, we are interested in the process that takes an image in its raw pixel form as input and generates a sequence of tokens that capture its visual details. The structure of the sequence from coarse to fine details is achieved by ordering the information starting with the most significant bits of the most significant wavelet coefficients. En par with wavelet coding, we also have a goal to create token sequences that are as short as possible. This creates shorter contexts for the transformer decoder and improves its performance. As we shall see (Subsection \ref{subsec:voc}) it is quite easy to convert sequences of few wavelet tokens to shorter sequences at the tradeoff of using a larger vocabulary of tokens. 
\subsection{Wavelet Transforms}

 A univariate wavelet system \cite{Daubechies,Mallat} is a family of real functions in $L_2 ({\dsR})$ of the form $\left\{ {\psi_{j,k}  \thinspace  \colon \thinspace 
(j,k)\in {\dsZ}^2} \right\}$  built by dilating and 
translating a unique mother wavelet function $\psi$
\[
\psi_{j,k}(x) :=2^{-j/2}\psi (2^{-j}x-k),
\]
where the mother wavelet typically has compact support (or fast decay) and has $r$ vanishing moments

\begin{equation} \label{vanish}
\int_{\dsR} {x^k\psi (x)dx=0},\qquad k=0,1\ldots,r-1.
\end{equation} 

Wavelet systems can be constructed to serve a basis of $L_2(\dsR)$. To facilitate applications, one then also constructs a dual $\tilde{{\psi }}$ of $\psi $, where $\langle \psi_{j,k},\tilde{\psi}_{j',k'} \rangle=\delta_{j,j'}\delta_{k,k'}$, so that for each $f\in L_2(\dsR)$,  
\[
f=\sum\limits_{j,k} {\langle f,\tilde{{\psi }}_{j,k} \rangle \psi_{j,k} } .
\]
For special choices of $\psi $ 
, the set $\left\{ {\psi_{j,k} } \right\}$ forms an orthonormal basis 
for $L_2 ({\dsR})$ and then, $\psi =\tilde{{\psi }}$. 

Usually, one starts the construction of a wavelet system from a \textbf{\textit{Multi-Resolution Analysis (MRA)}} generated by a scaling function
$\varphi \in L_2 \left( {\dsR} \right)$ that satisfies a two-scale equation 
\[
\varphi =\sum\limits_k {a_k \varphi \left( {2\cdot -k} \right)} .
\]
One then sets 
\[
V_j =\overline {span} \left\{ 
{\varphi_{j,k} :=2^{{-j} \mathord{\left/ {\vphantom {{-j} 2}} \right. 
\kern-\nulldelimiterspace} 2}\varphi \left( {2^{-j}\cdot -k} \right):k\in 
{\dsZ}} \right\},\quad j\in \dsZ,
\]
which implies (under certain mild conditions)
\[
....V_2 \subset V_1 \subset V_0 \subset V_{-1} \subset V_{-2} ...,
\quad
\cap V_j =\left\{ 0 \right\},
\quad
\cup_j V_j =L_2 ({\dsR}).
\]
 Again, to facilitate applications, one may also construct a dual $\tilde{{\varphi}}$ of $\varphi$, where $\langle \varphi_{0,k},\tilde{\varphi}_{0,k'} \rangle=\delta_{k,k'}$, so that for each $f\in V_j$,  
\[
f=\sum\limits_{k} {\langle f,\tilde{{\varphi }}_{j,k} \rangle \varphi_{j,k} } .
\]
Equipped with the MRA, one then proceeds to construct the wavelet $\psi$ such that $W_j :=\overline {span} 
\left\{ {\psi_{j,k} :k\in {\dsZ}} \right\}$ with $V_{j+1} + W_{j+1} 
=V_j $. A classic example for an orthonormal MRA and wavelet system where $\varphi=\tilde{\varphi}$ and $\psi=\tilde{\psi}$, are the Haar scaling 
function and Haar wavelet
\[
\varphi \left( x \right):=\left\{ {{\begin{array}{*{20}c}
 {1,} \hfill & {x\in \left[ {0,1} \right],} \hfill \\
 {0,} \hfill & {\mbox{else}.} \hfill \\
\end{array} }} \right.
\quad
\psi \left( x \right):=\left\{ {{\begin{array}{*{20}c}
 1 \hfill & {x\in \left[ {0,\frac{1}{2}} \right),} \hfill \\
 {-1} \hfill & {x\in \left[ {\frac{1}{2},1} \right],} \hfill \\
 0 \hfill & \mbox{else.} \hfill \\
\end{array} }} \right.
\]

\begin{figure}[htbp]
\centerline{\includegraphics[width=4in]{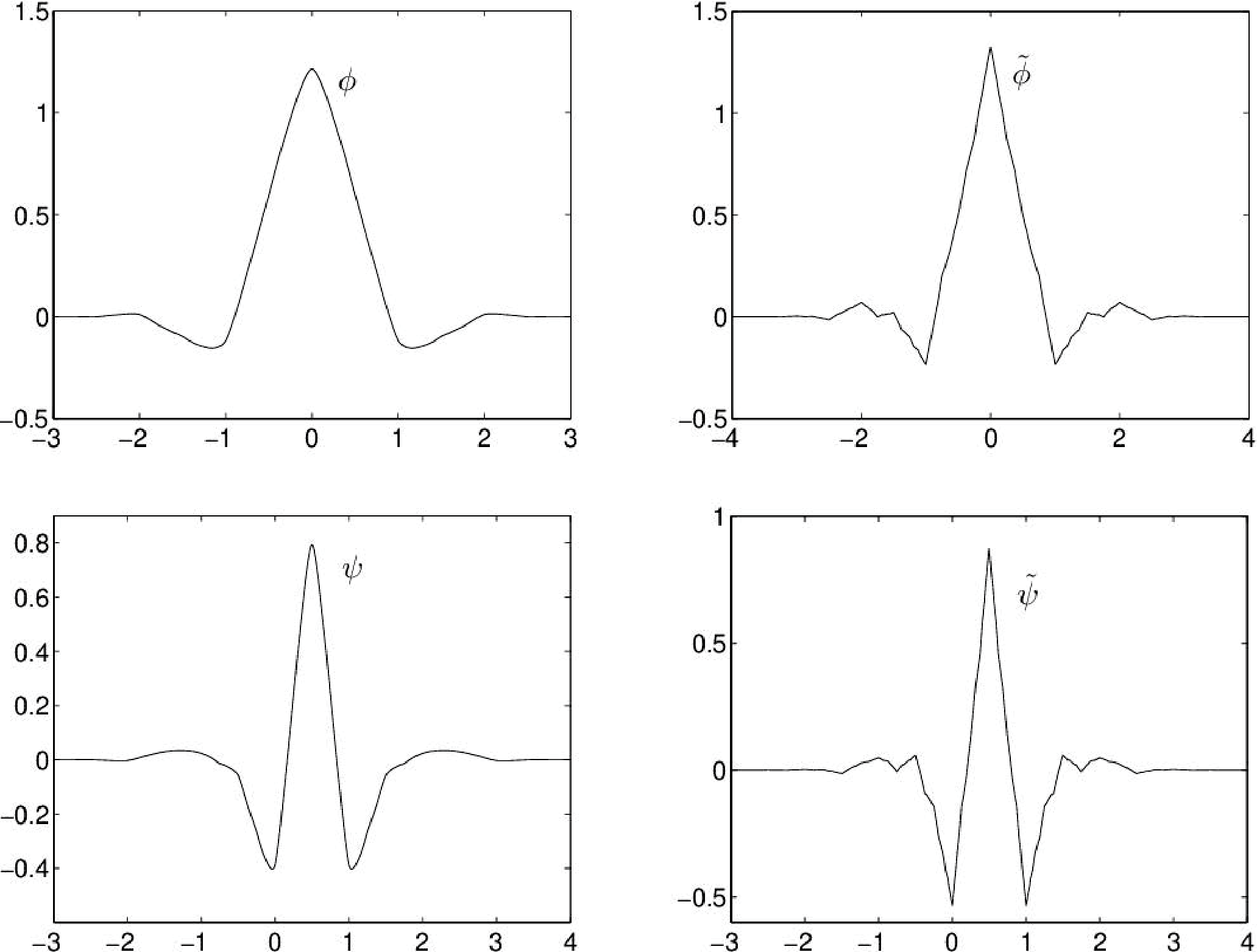}}
\caption{The CDF [9,7] wavelet system (figure reproduced from \cite{Cern2011DISCRETEC9}).}
\label{fig:97}
\end{figure}

The bivariate Haar system (see below) is a good choice when working with piecewise constant images, such as the MNIST handwritten digits~\cite{deng2012mnist}. For some of our experiments, we use a famous wavelet system from the Cohen Daubechies Feauveau (CDF) family of wavelets \cite{Daubechies}, which is sometimes termed bior4.4 ($r=4$ in \eqref{vanish}) or [9,7] in the signal processing community (the supports of the scaling functions and wavelets, as well as the lengths of the associated filters, are 9 and 7). The generating functions of the bior4.4 are depicted in Figure \ref{fig:97}.

The wavelet model can be easily generalized to any dimension, via a tensor product of the wavelet and the 
scaling functions. Assume that the univariate dual scaling functions $\varphi,\tilde{\varphi}$ and dual wavelets $\psi,\tilde{\psi}$, are given. Then, a wavelet bivariate basis is 
constructed using three types of basic wavelets
\[
\psi^1(x_1 ,x_2 ):=\varphi (x_1 )\psi (x_2 ),
\quad
\psi^2(x_1 ,x_2 ):=\psi (x_1 )\varphi (x_2 ),
\quad
\psi^3(x_1 ,x_2 ):=\psi (x_1 )\psi (x_2 ),
\]
\[
\tilde{\psi}^1(x_1 ,x_2 ):=\tilde{\varphi} (x_1 )\tilde{\psi}(x_2 ),
\quad
\tilde{\psi}^2(x_1 ,x_2 ):=\tilde{\psi}(x_1 )\tilde{\varphi}(x_2 ),
\quad
\tilde{\psi}^3(x_1 ,x_2 ):=\tilde{\psi}(x_1 )\tilde{\psi}(x_2 ).
\]
The bivariate wavelet transform of $f\in L_2({\dsR}^2)$, 
in terms of the bivariate wavelet tensor basis
\[
\psi^e_{j,k}:=2^{-j}\psi^e(2^{-j}\cdot-k), \quad
\tilde{\psi}^e_{j,k}:=2^{-j}\tilde{\psi}^e(2^{-j}\cdot-k),
\quad e=1,2,3, j\in\dsZ, k\in\dsZ^2,
\]
is then 
\[
f=\sum_{e=1,2,3,j\in\dsZ,k\in\dsZ^2} {\langle f, \tilde{\psi}^e_{j,k} \rangle \psi_{j,k}^e }.
\]
The bivariate wavelet decomposition can thus be interpreted as a 
signal decomposition in a set of three spatially oriented frequency 
subbands: $LH(e=1)$ detects horizontal edges; $HL$ ($e=2)$ detects 
vertical edges and $HH$ ($e=3)$ detects diagonal edges. 

\begin{figure}[htbp]
\centerline{\includegraphics[width=2.5in,height=2.4in]{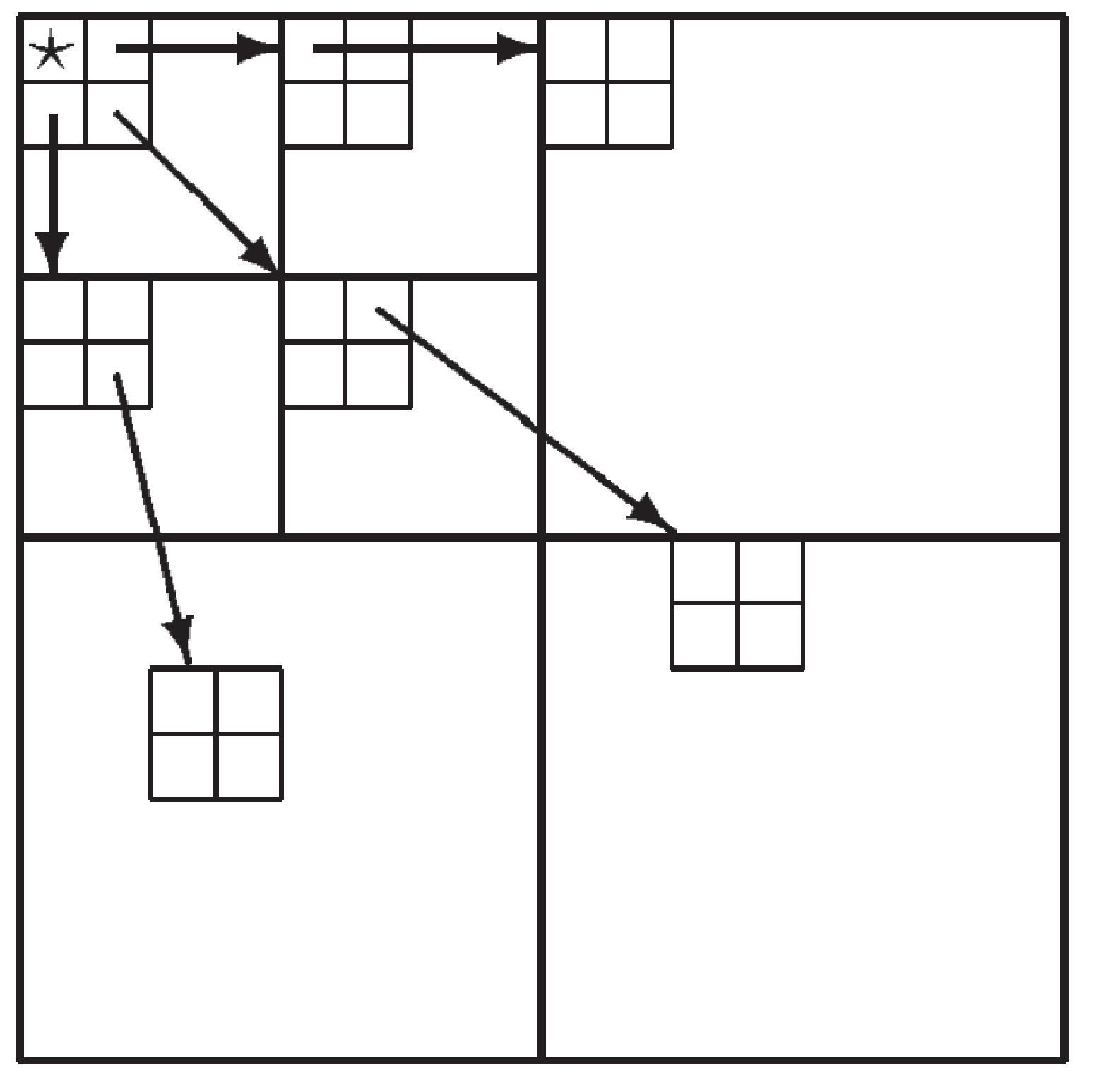}}
\caption{Wavelet coefficient tree structure across the subbands (MRA decomposition).}
\label{fig:quad}
\end{figure}

Under the assumption that $\psi$ and $\tilde{\psi}$ are compactly supported (or have fast decay), a wavelet coefficient $\langle f, \tilde{\psi}^e_{j,k} \rangle$ at a scale $j$ represents the information 
about the function in the spatial region of radius $\sim 2^j$ in the neighborhood of 
$2^jk,\thinspace k\in {\dsZ}^2$. At the next finer scale $j-1$, the 
information about this region is represented by four wavelet coefficients, 
which are described as the children of $\langle f, \tilde{\psi}^e_{j,k} \rangle$. This leads to a natural tree structure organized in a quad tree structure 
of each of the three subband types as shown in Figure \ref{fig:quad}. As $j$ decreases, the child 
coefficients add finer and finer details into the spatial regions occupied by their ancestors.

In image processing, one uses the Discrete Wavelet Transform (DWT). It works by initially assuming that the image pixels $\{f_k=f_{k_1,k_2}\}_{k1,k2=1}^M$ are good approximants of the projections on the shifts of the dual scaling function with the underlying function $f$ (see \cite[Section 7.3.1]{Mallat} for a detailed justification)
\[
f_k \approx \langle f,\tilde{\varphi}_{0,k} \rangle.
\]
With these coefficients as input, one uses the DWT to compute coefficients down to some predefined low-resolution $m$. For simplicity, we may assume that $M=2^m$ and that we use the DWT to compute
\begin{equation} \label{imageDWT}
\{\langle f,\tilde{\varphi}_{m,k} \rangle\}, \quad
\{\langle f,\tilde{\psi}^e_{j,k} \rangle\}, \quad 1\le j \le m,\quad e=1,2,3.
\end{equation}

\begin{figure}[htbp]
\centering
\subfigure[Fishing boat image.]{\includegraphics[width=0.3\textwidth]{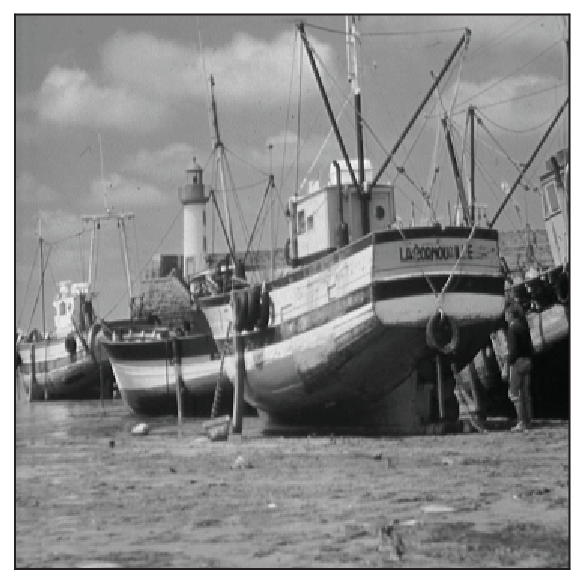}}
\hfill
\subfigure[15267 significant coefficients.]{\includegraphics[width=0.3\textwidth]{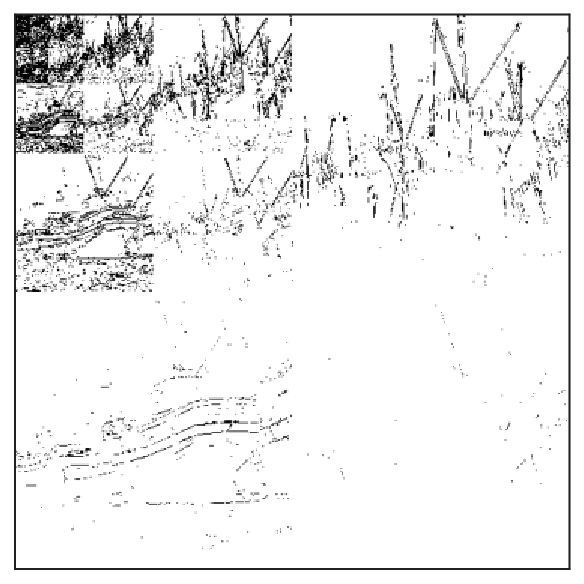}}
\hfill
\subfigure[Compressed image 1:17.]{\includegraphics[width=0.3\textwidth]{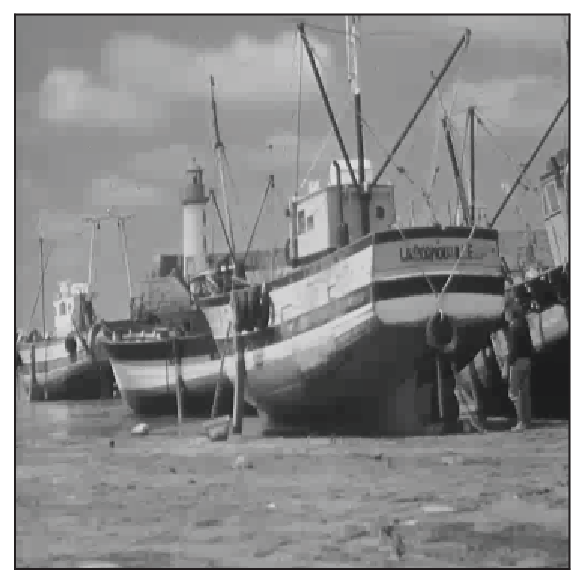}}
\caption{Image compression based on sparse wavelet approximation.}
\label{fig:BoatCompress}
\end{figure}

Wavelet representations are considered very efficient for image compression \cite{EZW, SPIHT, JPEG2000Book}. The edge information typically constitutes a small portion of a typical image, while the dual wavelet coefficients have a large absolute value only if edges intersect the support of the corresponding dual wavelets. Consequently, the image can be approximated well using a few significant 
wavelet coefficients. A clear statistical structure also follows: large/small values of wavelet coefficients tend to propagate through the 
scales of the quadtrees depicted in Figure \ref{fig:quad}. As an example, a sparse wavelet representation of a $512\times 512$ 
fishing boat image and a compressed version of it are shown in Figure \ref{fig:BoatCompress}, where the compression algorithm JPEG2000 is based on the sparse representation. The Figure clearly depicts that the significant wavelet coefficients (coefficients with relatively large absolute values) are located on strong edges of the image.

\subsection{Embedded Wavelet Tokenization} \label{subsec:token}

The sparse wavelet representation \eqref{imageDWT} of an image provides the perfect infrastructure for the generation of embedded coding representations \cite{EZW,SPIHT,JPEG2000Book}. Embedded coding is similar in spirit to binary finite precision representations of real numbers, where the ``encoding'' can cease at any time and provide the ``best'' approximation of the real number achievable within the framework of the binary digit representation. Similarly, the embedded coder can cease at any time and provide the ``best'' representation of an image achievable
within its framework. Embedded coding streams can be generated from wavelet representations by ordering the information on the wavelet representation starting with the most significant bits of the most significant coefficients. That is, the coefficients with the largest absolute value. In image coding applications the goal is to generate a compressed bit stream of `0' and `1's. This can be efficiently achieved by using information theoretical tools such as arithmetic coding. Here, our goal is somewhat different where we aim to create an efficient tokenization method conforming to the following two objectives:
\begin{enumerate}
  \item [(i)] Ensuring statistically frequent structural patterns - The existence of common patterns within the token sequence allows the language models to learn them as contexts when they attempt to generate the most probable next token in the context of the previous tokens. The wavelet tokenization processes described below provide that by creating token sequences that are ordered based on coefficient absolute value and then resolution. It is known \cite{EZW,SPIHT}, that there are strong correlations between insignificant coefficients with their `ancestors' at lower resolutions 
(see Subsection \ref{subsec:zt}).
  \item [(ii)] Trade-off between sequence length versus number of tokens - When token sequences become very long, the LLMs need to deal with longer contexts, which can be challenging. At the same time, a dataset of possibly shorter sequences that are based on a large vocabulary of tokens can also be challenging for different reasons. As we shall see, the method of Subsection \ref{subsub:encode} uses 7 tokens and may create long sequences. The method of \ref{subsec:zt} is a somewhat more advanced and manages to both reduce the number of tokens by 1 and at the same time reduce the sequence lengths significantly as the dimensions of the images increase. In Subsection \ref{subsec:voc} we review standard methods that allow to control this tradeoff by merging frequent sub-sequences into new tokens. 
\end{enumerate}

First, for simplicity of notation, using \eqref{imageDWT}, denote for $I=(i_1,i_2)$, $1\le i_1,i_2 \le 2$
\[
\alpha_I := \langle f,\tilde{\varphi}_{m,I} \rangle.
\]
We also map the coefficients $\{\langle f,\tilde{\psi}^e_{j,k} \rangle\}$, $1 \le j \le m$, $e=1,2,3$,  
\[
\alpha_I \leftarrow \langle f,\tilde{\psi}^e_{j,k} \rangle,
\]
based on their location 
\[
I=(i_1,i_2), \quad i_1 \ge 3\vee i_2 \ge 3, \quad i_1\le M \wedge i_2 \le M,
\]
in the coefficient matrix. We note in passing that one may assume that the low resolution scaling function coefficients from  \eqref{imageDWT} are known during training and are randomly sampled from some distribution during image generation and therefore need not be part of the tokenization. 

\subsubsection{Encoding a wavelet representation into a token sequence} \label{subsub:encode}

We now show how to process the numeric representations of the coefficients, from most significant to least significant and `encode' them into a relatively compact series of tokens. The representation using the series of tokens should be invertible. That is, one should be able to convert (e.g. `decode') the token sequence back to the wavelet representation. 

To this end, assuming the image pixels are normalized to the range $[0,1]$, one can show that for an image of dyadic dimension $[M,M]=[2^m,2^m]$, after $m-1$ iterations of the bivariate DWT
\begin{equation} \label{coefbound}
\max_{I}|\alpha_{I}| \le 2^{m-1}.
\end{equation}
Assuming for simplicity that all images of a given dataset have the same dyadic dimensions $[M,M]$, then this bound holds for all of their wavelet representations. Our first option is to initialize a threshold $T = 2^{m-2}$ and begin scanning the wavelet coefficients of the image, in a predetermined order (see below) for significance, with the goal of reporting only those coefficients for which the following holds
\[
T \le |\alpha_{I}| < 2T.  
\]
Our second option, is to compute separately for each image in the dataset 
\begin{equation} \label{coefbound2}
\tilde{m}:=\lceil{\log_2 \max_{I}|\alpha_{I}|}\rceil, 
\end{equation}
and then initialize for the specific image $T=2^{\tilde{m}-1}$. In this scenario, we store and use the parameter $\tilde{m}$ for each image in the training set along with its sequence of tokens. 

We also maintain a matrix of approximated wavelet coefficients $\{\tilde{\alpha}_{I}\}$ which we initialize with zeros. Once we complete the processing at a given bit plane, we update $T \leftarrow  T/2$ and repeat the process. At each bit-plane we report the significant coefficients that were just uncovered in this bit-plane using a token `NowSignificantNeg' if the coefficient is negative or a token `NowSignificantPos' if it is positive. At the time of uncovering, we modify the approximation of the coefficient $\tilde{\alpha}_{I}$ to have the absolute value $3T/2$, with the reported sign. Next, we add a token to represent the coefficient's next significant bit, `NextAccuracy0' if the coefficient satisfies $|\alpha_{I}|\le |\tilde{\alpha}_{I}|$ or the token `NextAccuracy1' if $|\alpha_{I}|> |\tilde{\alpha}_{I}|$. The approximation $\tilde{\alpha}_{I}$ is updated accordingly by subtracting or adding $T/4$ (depending on the sign of the coefficient and the accuracy bit type).

Let us demonstrate with an example. Assume T=16 and $\alpha_{I}=-17.45$. Therefore, the coefficient is first uncovered in the current bit plane. When we arrive at the index $I$, we report a `NowSignificantNeg' token for this coefficient, providing it with a temporary approximation $\tilde{\alpha}_{I}=-24$, which lies in the middle of the segment $[-T,-2T]=[-16,-32]$. Next, since in fact $|\alpha_{I}|\le 24$, we report a token `NextAccuracy0' to represent the coefficient's next significant bit, providing an updated approximation $\tilde{\alpha}_{I}=-20$, which lies in the middle of the segment $[-T,-3T/2]=[-16,-24]$, leading to a better approximation of the ground truth value. 

In case a coefficient has been uncovered in any of the previous bit-planes and is already known to be significant, we only add one of the tokens `NextAccuracy0' if $|\alpha_{I}|\le |\tilde{\alpha}_{I}|$ or `NextAccuracy1' if $|\alpha_{I}| > |\tilde{\alpha}_{I}|$. We then update the approximation $\tilde{\alpha}_{I}$ by subtracting or adding $T/4$ (depending on the sign of the coefficient and the accuracy bit type).

Assuming the bit-plane scanning order of the coefficients is fixed, one then only needs to add the token `Insignificant' to provide a valid invertible tokenization process. One  simply scans the coefficients in the fixed order and uses their true known value to test and apply one of three possibilities:
\begin{enumerate}
    \item[(i)] $|\alpha_{I}| \ge 2T$: The  coefficient has already been reported as significant in a previous bit-plane. Therefore one reports the token `NextAccuracy0' or `NextAccuracy1' depending on the test $|\alpha_{I}| < |\tilde{\alpha}_{I}|$.
    \item[(ii)] $T \le |\alpha_{I}| < 2T$: First report the token `NowSignificantNeg' or `NowSignificantPos' depending on the sign and then report the token `NextAccuracy0' or `NextAccuracy1'.
    \item[(iii)] $|\alpha_{I}|<T$: report `Insignificant'.  
\end{enumerate} 

The process described above, although completely sufficient for invertible tokenization, potentially creates long sequences. Specifically, it does not take into consideration the local correlations among `neighboring' insignificant wavelet coefficients. Due to the sparsity property of the wavelet transform, during the scanning process, many of the `Insignificant' coefficients form local groups. Moreover, there are correlations between local groups of insignificant coefficients of the same subband type across resolutions in the manner of the quad-tree structure of Figure \ref{fig:quad}. Image compression algorithms such as the EZW \cite{EZW} or SPIHT \cite{SPIHT}, are based on statistical zero tree models that try to capture these correlations across resolutions (see the Zero-Tree method in the next subsection). As we shall later see, for image generation, we actually rely on the powerful capabilities of the transformer models to learn correlation patterns of the `wavelet language' of the given dataset. However, we do `ease the burden' off the transformers significantly by utilizing the structure of the groups of insignificant coefficients to reduce the size of the token sequences, thereby creating shorter contexts. 

To this end, we add two additional tokens for groups of insignificant coefficients: `Group4x4' and `Group2x2' and modify the scanning process to visit the coefficients based on groups of $4\times 4$. The first token is used in locations where the scan is at an index $(4l_1,4l_2)$, for some integers $l_1,l_2$. If at the current bit plane, all the 16 coefficients with indices $I=(i_1,i_2)$, $4l_1\le i_1 \le 4(l_1+1), 4l_2\le i_2\le 4(l_2+1)$, are still insignificant, we issue the token `Group4x4' and the tokenization process continues to the next group of $4\times 4$ coefficients. However, if any of the coefficients of the $4\times 4$ group becomes significant in the current bit-plane, the group breaks down to 4 groups of $2\times 2$. If a group of $2 \times 2$ is still composed of insignificant coefficients at the current bit-plane, we add a token `Group2x2'. If a group of $2\times 2$ breaks down, then each coefficient from the group is reported individually as being `Insignificant' or one of `NowSignificantNeg', `NowSignificantPos'. The scanning process keeps track of which groups broke up, so that only necessary and informative tokens are generated. We summarize the seven tokens and their roles below

\begin{enumerate}
    \item [(i)] `Group4x4' -- At the index $(4l_1,4l_2)$, the group of 16 coefficients $\{\alpha_{I}\}$, $4l_1\le i_1\le 4l_1+4$, $4l_2\le i_2\le 4l_2+4$, are still insignificant, $|\alpha_{I}|<T$.
    \item [(ii)] `Group2x2' -- At the index $(2l_1,2l_2)$, the group of 4 coefficients $\{\alpha_{I}\}$, $2l_1\le i_1\le 2l_1+2$, $2l_2\le i_1\le 2l_2+2$, are still insignificant, $|\alpha_{I}|<T$.
    \item [(iii)] 'NowSignificantNeg', 'NowSignificantPos' -- At the current location $I$, the coefficient satisfies $T\le|\alpha_{I}|<2T$. If the coefficient was part of a group of insignificant coefficients at the previous bit-plane,  the group is now automatically dissolved. 
    \item [(iv)] `Insignificant' -- At the current location $I$, the coefficient is still insignificant and satisfies $|\alpha_{I}|<T$. If the coefficient was part of a group of insignificant coefficients at the previous bit-plane, the group is now automatically dissolved.
    \item [v)] `NextAccuracy0', `NextAccuracy1' -- At the current location $I$, the coefficient has already been reported to be significant since it satisfies $|\alpha_{I}|\ge T$. Here, we improve the accuracy of its approximation using one of these tokens, depending on the test $|\alpha_{I}| < |\tilde{\alpha}_{I}|$.
\end{enumerate}

\begin{figure}[H]
\centering
\subfigure[Outer subband scanning order.]{\includegraphics[width=0.3\textwidth]{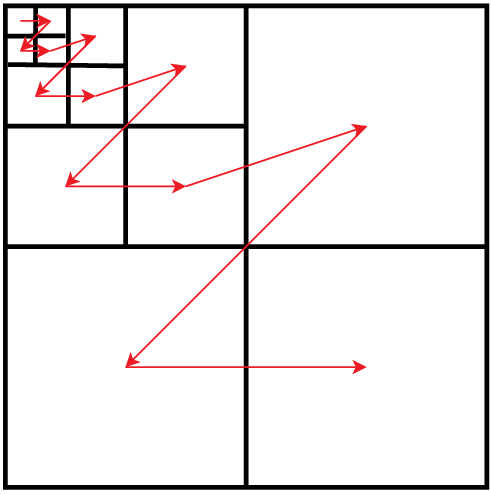}}
\hspace{1cm}
\subfigure[Inner scanning order of $4\times 4$ blocks.]{\includegraphics[width=0.3\textwidth]{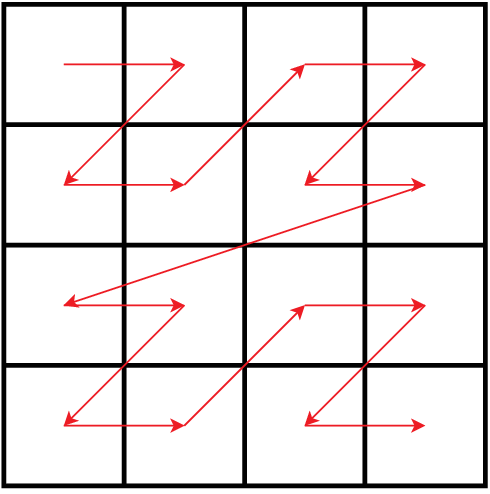}}
\caption{A sketch illustrating the outer and inner scanning orders.}
\label{fig:scanning}
\end{figure}

The bit-plane scan is carried out in two nested loops; the outer loop proceeds from low resolution to high resolution, each time traversing the three types of wavelet subbands. The inner loop traverses the $4\times 4$ blocks. Figure~\ref{fig:scanning} illustrates the outer and inner scanning patterns. 

\begin{figure}
\centering
\subfigure[$32\times 32$ padded MNIST image.]{ \label{fig:padded_MNIST_image}\includegraphics[width=0.3\textwidth]{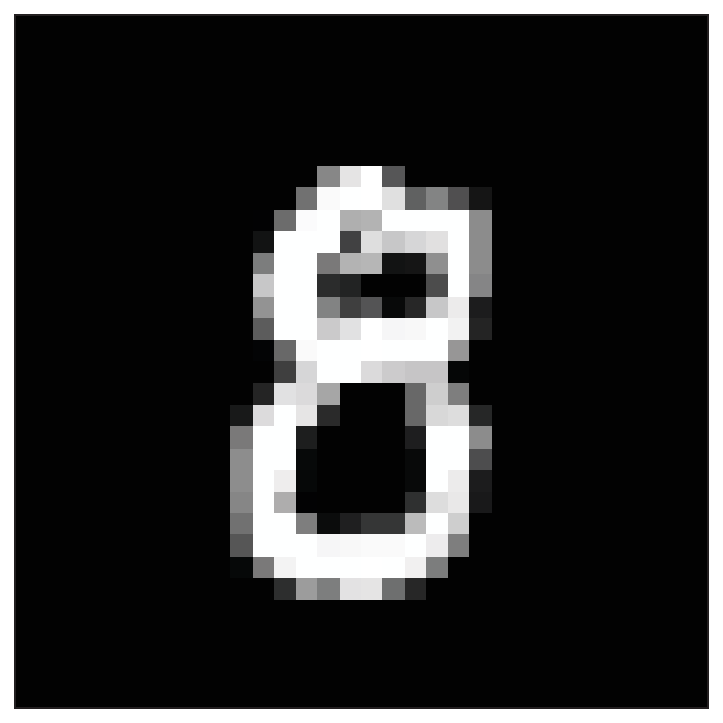}}
\hfill
\subfigure[Wavelet transform.]{\label{fig:wavelet_transform}\includegraphics[width=0.3\textwidth]{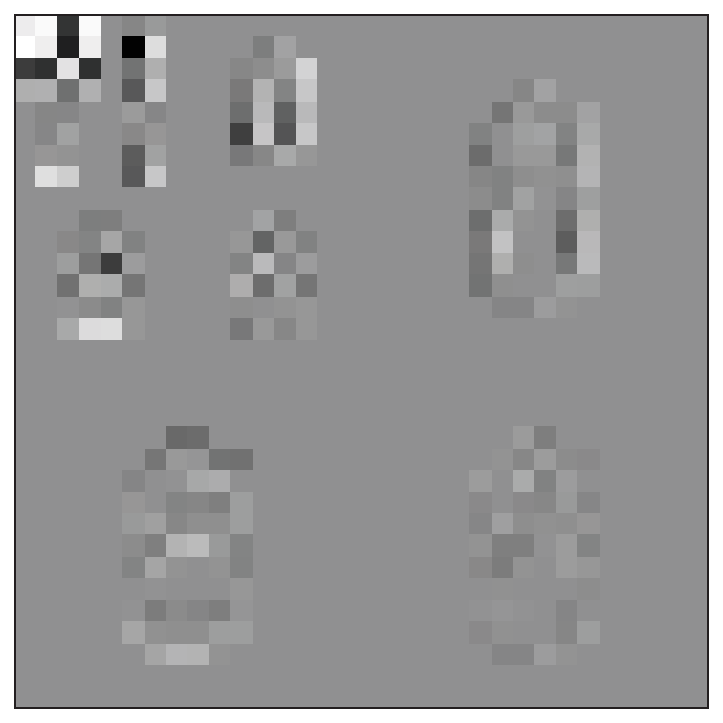}}
\hfill
\subfigure[Significant coefficients after 3 bit-planes.]{ \label{fig:three_bit_planes}\includegraphics[width=0.3\textwidth]{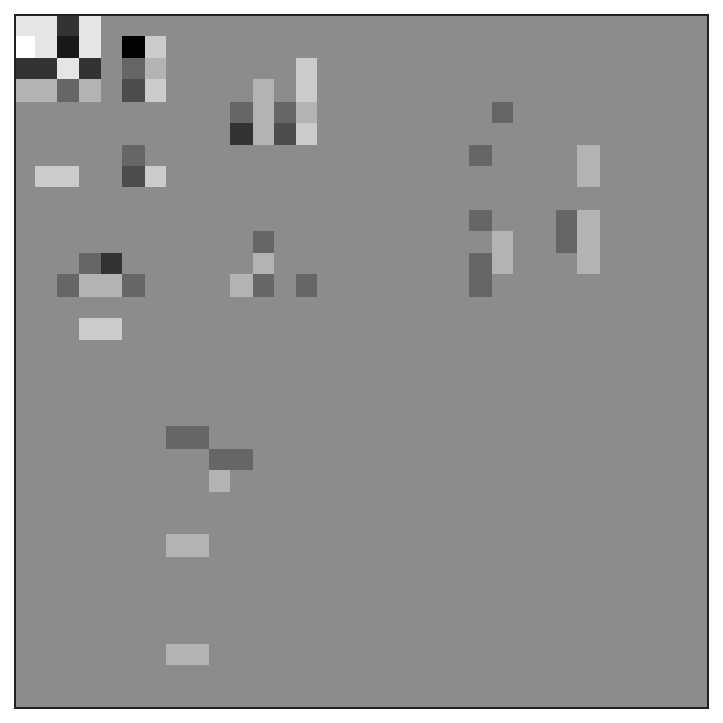}}
\vskip\baselineskip
\subfigure[Wavelet domain tokenization - first bit plane.]{\label{fig:first_bit_plane}\includegraphics[width=0.3\textwidth]{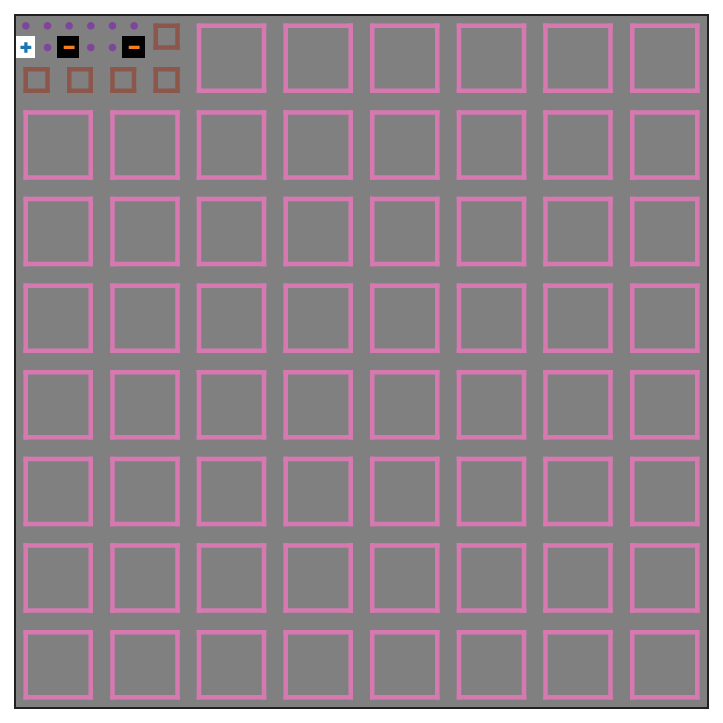}}
\hfill
\subfigure[Wavelet domain tokenization - second bit plane.]{\label{fig:second_bit_plane}\includegraphics[width=0.3\textwidth]{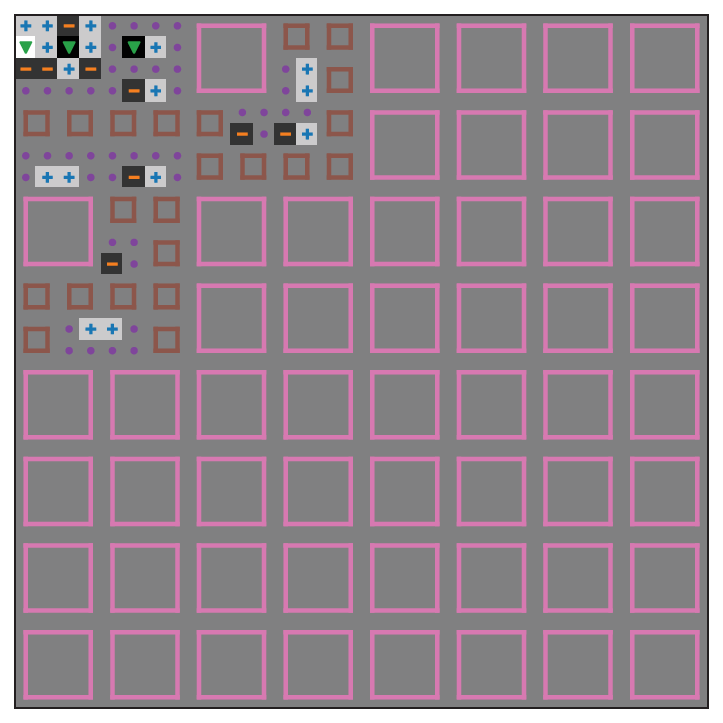}}
\hfill
\subfigure[Wavelet domain tokenization - third bit plane.]{\label{fig:third_bit_plane}\includegraphics[width=0.3\textwidth]{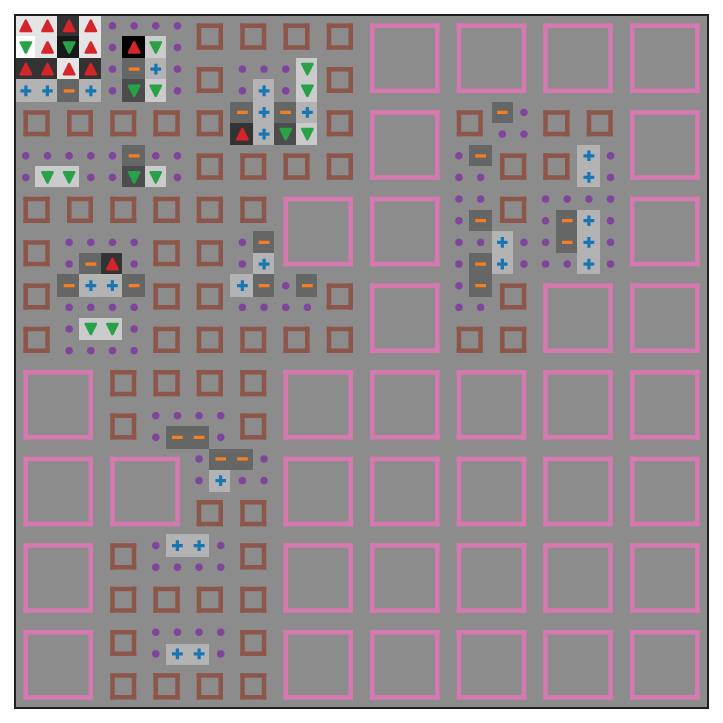}}
\caption{Depiction of the tokenization process. On the top left and middle, a $32\times 32$ padded MNIST image and its wavelet transform. On the top right, the wavelet approximation generated by the first three bit-planes. The bottom row illustrates the tokens and their locations on the $32\times 32$ grid, where, `NowSignificantNeg' and `NowSignificantPos' tokens are annotated with orange ``$-$'' and blue ``$+$'' and signs respectively. The tokens `NextAcurracy0' and `NextAccuracy1' are marked with green down and red up triangles. The purple dots represent `Insignificant' coefficients and the brown and pink squares represent the `Group2x2' and `Group4x4' zero block tokens.}
\label{fig:tokenization_rounds}
\end{figure}

Figure~\ref{fig:tokenization_rounds} exemplifies the tokenization algorithm of an image from the MNIST dataset~\cite{deng2012mnist}. The image was padded with zeros to be of dimensions $M\times M = 32\times 32$, with $m=5$. The bottom row of the figure shows the tokens and their locations on the wavelet image for the first three bit planes. To make the process clearer, we explicitly write the resulted sequence of tokens for the first bit plane shown in Figure~\ref{fig:first_bit_plane}.
\begin{align*}
    \{ &\text{`Insignificant', `Insignificant', `NowSignificantPos', `Insignificant', }
    \\ & \text{`Insignificant', `Insignificant', `NowSignificantNeg', `Insignificant', }
    \\ & \text{`Group2x2', `Group2x2', }
    \\ & \text{`Insignificant', `Insignificant', `Insignificant', `NowSignificantNeg', }
    \\ & \text{`Group2x2',`Group2x2', `Group2x2',}
    \\ & \text{`Group4x4'}, \dots, \text{`Group4x4'}\}
\end{align*}
The token sequences of the second and third bit-planes follow the same scanning pattern. Eventually, the three sequences are concatenated in the natural order to form the final sequence which describes the three bit planes wavelet image appearing in Figure~\ref{fig:three_bit_planes}. 

There is a very important hyper-parameter which is the choice of the smallest threshold at the final bit-plane. Through this hyper-parameter, the wavelet representation provides us with a very robust and stable trade-off of fine detail generation and length of token sequences. Choosing a final threshold provides very consistent control over visual quality relating to: ``Visually Lossless'', ``High'', ``Medium'', ``Low'', etc. This is en par with the quality settings in digital cameras, which in turn lead to a selection of the corresponding quantization tables of the JPEG algorithm generating the compressed images. 

\subsubsection{Zero-tree tokenization} \label{subsec:zt}

The Zero-Tree method is an alternative tokenization method that is aligned with \cite{EZW} and provides shorter sequences, especially as image size increases. The zero-tree approach leverages on the correlations of insignificant coefficients across resolutions. Statistically, if a wavelet coefficient at some resolution is insignificant, then with very high probability (around $90\%$ for real life images) its descendants at the same subband and higher resolutions will also be insignificant. With the zero-tree tokenization method the scanning visits coefficients from low to high resolution and the tokens `Group2x2' and `Group4x4' are replaced with a single `zero-tree' token. If the token is reported at a certain location in the scan, then it is understood that the coefficient at this location as well as all its descendants are still insignificant at the current bit-plane. The descendants of a coefficient at 
location 
\[
I=(i_1,i_2), \quad i_1 \ge 3\vee i_2 \ge 3, \quad i_1\le M/2 \wedge i_2 \le M/2,
\]
are its children 
\[
\{(2i_1,2i_2),(2i_1,2i_2+1),(2i_1+1,2i_2),(2i_1+1,2i_2+1)\}
\]
and then recursively their children. Once a coefficient is reported as a `zero tree' coefficient, it is understood that all of its descendants are still insignificant and the scanning skips them.
For the FashionMNIST dataset (see examples below) the mean token sequence length is 1822.5 for the zero blocks tokenization method 
and 1601.7 for the zero tree method, although the latter uses 6 tokens instead of 7.

\subsubsection{The trade-off of vocabulary size and sequence lengths} \label{subsec:voc}

\begin{figure}[htbp]
\centerline{\includegraphics[width=3in]{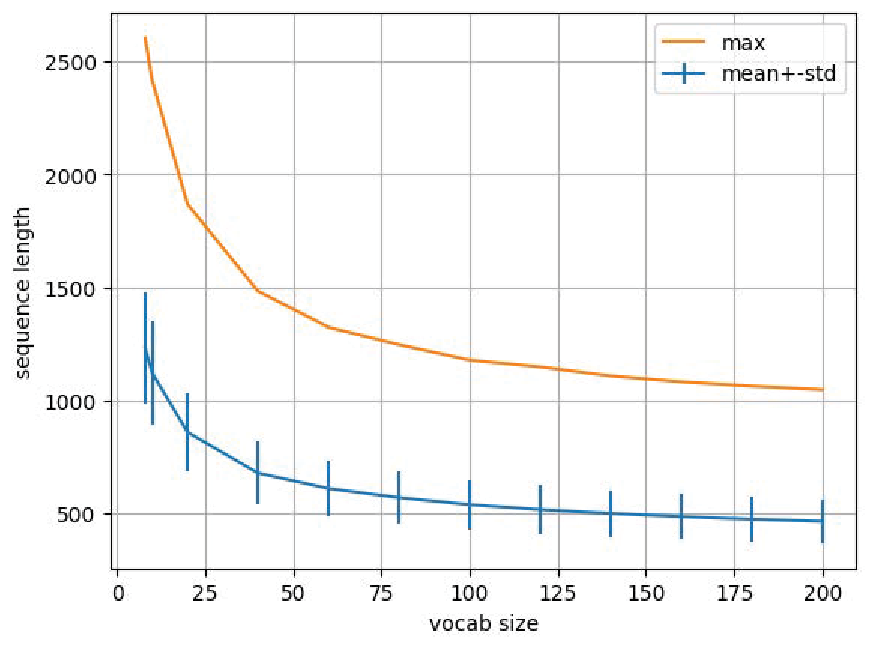}}
\caption{The trade-off between vocabulary size and token sequence length for the fashionMNIST dataset.}
\label{fig:vocab}
\end{figure}

It is quite standard in the field of autoregressive methods to control the tradeoff between the token vocabulary size and the dataset's mean token sequence length in an attempt to find
the optimal configuration for given computational resources and model architecture. Since the wavelet method uses a relatively very small number of tokens (language models typically support a vocabulary of tens of thousands of tokens) and creates relatively long sequences, it is relevant in scenarios where the computational resources do not allow the use of long token sequences. One of the simplest methods is Byte Pair Encoding (BPE) \cite{BPE}. BPE is a subword tokenization technique commonly used in natural language processing, which iteratively merges the most frequent pairs of tokens into new tokens, thereby creating a more compact representation of the data. We used HuggingFace's tokenizers library \cite{hugtoken} to generate Figure \ref{fig:vocab}

\subsubsection{Decoding the token sequence into an approximate wavelet representation}

The tokenization process described in the previous subsections can be easily inverted back to an approximate wavelet representation. Moreover, any initial sub-sequence can be inverted to provide a possibly coarser approximation. We initialize a matrix of size $M\times M$ of the approximated wavelet coefficients $\{ \tilde{\alpha}_{I} \}$ with zeros and begin the scanning process with the first bit-plane. Based on \eqref{coefbound} or \eqref{coefbound2}, we know how to initialize the first bit-plane with the initial threshold $T=2^{m-2}$ or $T=2^{\tilde{m}-1}$. We then process the token sequence and update the approximated coefficients using the corresponding `significant' and `bit accuracy' tokens. If for any given reason, the sequence of tokens terminates, we have the best possible approximated coefficients  $\{\tilde{\alpha}_{I}\}$ from which we can obtain an approximated image by applying the inverse DWT. Our decoding process relies on the assumption that the token sequence is valid. For example, a `Group4x4' token cannot appear while the decoder scan position is at a location of indices not divisible by 4. It is obvious how to achieve this in the context of image coding. However, during an image generation process, this needs to be enforced using the conditional next-token inference described in Subsection \ref{subsec:condtoken}.   

\section{The Generative Wavelet Transformer} \label{sec:GWT}

Assume that for a certain dataset of images, we have established the translation of the visual information of each image to a sequence of tokens encapsulating the visual information from coarse to fine details as explained in Subsection \ref{subsec:token}. We assume that within the sequences, distinct patterns and relations exist between the tokens. For example, the wavelet coefficients $\{\langle f,\tilde{\psi}^e_{j,k} \rangle\}$ of wavelets $\{\psi^e_{j,k}\}$ whose support intersects with a certain portion of an edge of the image, will be significant and aligned across scales in a tree-like structure as per Figures \ref{fig:quad} and \ref{fig:BoatCompress}(b). At the same time, coefficients of wavelets whose support intersects with a smooth area of the image will be insignificant and they appear in local groups. As explained, they also have a tree structure across scales, that can be captured explicitly by a `zero tree' token. This leads to the intuition that the powerful transformers created over the last few years \cite{FormalTrans} are able to learn the patterns of the `wavelet language' and to generate them from some random seeds during inference.

In this section, we describe how we modified the architecture of the DistilGPT2 transformer model \cite{sanh2019distilbert} to optimize it to align with the wavelet-based image generation method. This obviously requires training the modified model from scratch. We found it useful to use the code from HuggingFace \cite{distilGPT2} as a starting point. 

\subsection{Token vector representation} \label{subsec:vector}

Typically, in the standard scenarios of spoken languages, transformers apply a `pre-processing' learnable transform to tokens to convert them to vector representations. The idea is that similar words should be converted to vectors with some proximity, which intuitively serve as better input for the transformer's neural network. However, with the method of Subsection \ref{subsub:encode}, our wavelet dictionary includes only 7 tokens that have very distinctive and different roles. Therefore, the simple transformation of the tokens to the one-shot encoding of the standard basis of dimension 7 is probably a better, if not optimal choice. Thus, the initial vector representation of a token is: $\text{`Group4x4'}\rightarrow (1,0,0,0,0,0,0)$,  $\text{`Group2x2'}\rightarrow (0,1,0,0,0,0,0)$, etc. Therefore, in our `wavelet' transformers the `token $\rightarrow$ vector' learnable transformation is removed. 

\subsection{Initial bit-plane threshold} \label{subsec:thresh}

\begin{figure}[H]
    \centering
    \includegraphics[scale=0.6]{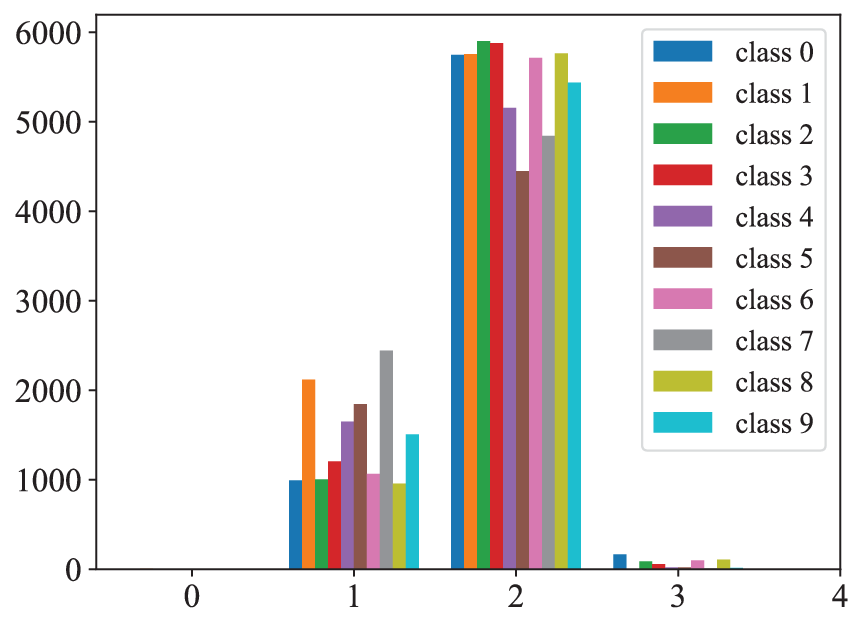}
    \caption{Distribution of $\log_2$ of the initial thresholds for the 70,000 MNIST images with the Haar wavelet transform.}
    \label{fig:MNIST_thresholds_distribution}
\end{figure}

Recall that we have two options: to use a uniform initial bit-plane threshold for all images in the dataset derived from \eqref{coefbound}, or to use an adaptive initial threshold for each image of the training set using \eqref{coefbound2}. In the latter case, we need to inform the transformer, per image, which initial threshold the token sequence is associated with. We do this as follows: assume a given dataset has $l$ possible values for $\tilde{m}$ in \eqref{coefbound2} (e.g., $l=4$ for the MNIST dataset, see Figure~\ref{fig:MNIST_thresholds_distribution}). Then, we concatenate a one-shot encoding of dimension $l$ of the initial threshold parameter of the given image to each vector representation of each token.

For image generation, one may sample randomly from the distribution of $l$ possible initial thresholds. In the case that the image generation is conditioned on a certain class (see Subsection \ref{subsec:guide}), one can sample from the conditional distribution of the possible thresholds of the specific class.  

\subsection{Positional encoding}

In classic transformer architectures \cite{FormalTrans}, one adds the positional encoding $v_p(t)$ of the position $t$ to the token's vector representation $v_e(x(t))$. Learnt positional embedding applied a learned transform $t\rightarrow v_p(t)$. Some transformers use hard-coded mapping of the position. Assuming the vector embedding dimension is $d_e$ and the maximum length of a sequence is $l_{\max}$, then 
\[
v_p(t)(2i-1)=\sin(t/l_{\max}^{2i/d_e}), \quad v_p(t)(2i)=\cos(t/l_{\max}^{2i/d_e}), \quad 1\le i\le d_e/2. 
\]
In our scenario of the wavelet language, the position of a token in a sequence is $(bp,I)$, where $bp$ is the enumeration of the bit-plane and $I=(i_1,i_2)$ is the index of the current coefficient $\alpha_I$ in the scan order. Therefore, we concatenate to the vector representation of a token from Subsection \ref{subsec:vector}, a vector component of dimension 3 with the location of the token $(bp,i_1,i_2)$. 

\subsection{Generative guidance} \label{subsec:guide} 

It is obviously critical for any image generation method to allow guidance of the generative process by placing a condition on the class type of the generated image or a text prompt that describes it. Some image generation  models apply a joint embedding space for text and images for this purpose. One such method is to used a pretrained model such as CLIP \cite{radford2021learning} that maps text and images to a joint embedding space. The CLIP contains an image encoder $f$ and a caption encoder $g$, that during training over pairs of images with captions $\{(x,c)\}$, optimizes a contrastive cross-entropy loss that encourages high dot-products $\langle f(x), g(c)\rangle$ in the joint embedding space. Thus, any image generation method, can use the vector embedding of the given text prompt $c$ to guide the generative process by conditioning the image embedding $f(x)$ to be highly correlated with the embedding of the textual prompt. 

In our case, since we converted the problem of image generation to a `wavelet-language' generation, we can apply `text'-type prompting methods. Having access to a joint embedding text-image space allows us to train using the vector representation of the image training set. Then, at image generation, we use the vector representation of the given text prompt to guide the generative process. There are very simple ways of using these vector representations. We choose to concatenate them to the vector representation of each token and its position (as explained above). For example, as shown in Section \ref{sec:experiment}, for the image datasets MNIST or FashionMNIST with 10 classes, it is easy to concatenate a vector of length 10 representing the class of the image. In the case where we wish to guide the generative process using a textual prompt, we may concatenate the CLIP vector embedding  \cite{radford2021learning} of the textual prompt to each token vector representation. As we discuss in Subsection \ref{subsec:blob}, we hope this approach to guiding the generative process can be generalized to composition of blobs \cite{nie2024compositional}, where a given guiding vector of a blob is used only at positions of the scan where the support of the corresponding wavelet intersects the blob.        

\subsection{Initialization of the generative process}

Since the guidance of the generative process (Subsection \ref{subsec:guide}) is applied through the concatenation of vector representations to each token vector representation, in some cases, the initialization becomes a minor issue. For example, when training on MNIST and generating digits, one can get away with a simple random choice from the subset: `Insignificant', `NowSignificantNeg' or `NowSignifiantPos' for the first token and from there the transformer will generate a valid token sequence which is converted to an adequate image of a digit from the pre-selected class. 

A more robust method is as follows. Suppose we wish to generate a handwritten digit from a certain digit class. Let $\{f_s\}_{s\in S}$ be the subset of MNIST images from that specific digit class and let 

\begin{equation} \label{classquads}
\{\langle f_s,\tilde{\varphi}_{m,k} \rangle\}, \quad s\in S, \quad k=(k_1,k_2), \quad 1\le k_1,k_2 \le 2,
\end{equation}
be the subset of low-resolution coefficients of these images defined by \eqref{imageDWT}. Let $N(v,\Sigma)$ be the fourth-dimensional normal distribution, approximated by the subset \eqref{classquads}. We then sample from $N(v,\Sigma)$, a random group of four low-resolution coefficients. Now, the token representation of these coefficients can serve as a basis for a robust initialization of the generative process of the required digit. In the case where the guidance is provided by a vector representation of some text-prompt, one can create the normal distribution using a subset of $K$-nearest neighbors in the image vector representation space.   

Once some random seeding allows us to initialize the token sequence, we may introduce as much diversity as required using the methods of Subsection \ref{subsec:control} so that even using the same seed may generate various images corresponding to the given guidance.   

\subsection{Conditional next token inference} \label{subsec:condtoken}

In Greedy generative mode, using the method of Subsection \ref{subsub:encode}, the next selected token $x(t)$, $1\le x(t) \le 7$, at location $t$, is the token for which the transformer assigns the highest probability from $(p_1(t),...,p_7(t))$.  As described in Subsection \ref{subsec:control} below, there are various alternative methods to control the output of the transformer. However, since each generative token inference step is a statistical event, it may occur that the next predicted token is not valid at the current position of the wavelet bit-plane scan. To overcome this, we apply conditional probability to 
ensure any selected token satisfies the conditions below relating to the context and the current position in the scan.    

\begin{enumerate}
    \item [(i)] `Group4x4' - The scan is at an index $(4l_1,4l_2)$ and the group has not yet dissolved. 
    \item [(ii)] `Group2x2' - The scan is at an index $(2l_1,2l_2)$ and the group has not yet dissolved. 
    \item [(iii)] 'NowSignificantNeg', 'NowSignificantPos' - At the current location $I$, the coefficient $\alpha_I$ is still insignificant, possibly as part of a group of insignificant coefficients. 
    \item [(iv)] `Insignificant' - At the current location $I$, the coefficient $\alpha_I$ is still insignificant, possibly as part of a group of insignificant coefficients.
    \item [v)] `NextAccuracy0', `NextAccuracy1' - At the current location $I$, the coefficient $\alpha_I$ has already been reported to be significant.
\end{enumerate}

\subsection{Controlling the degree of generative diversity during inference} \label{subsec:control}

Since we are applying a language transformer model we may use various simple stochastic mechanisms to control the generative process during inference and allow a diversity of possible images to be generated from a single prompt. Some of the available stochastic methods are: Beam search with multinomial sampling, Top-$k$ and Top-$p$. In our experiments, we tested the latter two:

\begin{enumerate} 
\item [(i)] Top-$k$ sampling - The Top-$k$ inference method \cite{fan-etal-2018-hierarchical} filters the $k$ most likely next words first and then samples from the probability mass that is redistributed among only those $k$ next words. GPT2 adopted this sampling scheme, which was one of the reasons for its success in story generation. In Figure \ref{fig:sandals} below, we see a diversity of sandals generated by guiding the model with the vector representation of the corresponding FashionMNIST `sandal' class and using the Top-$2$ method. We see that using $k=2$ is sufficient to move the generative process from a deterministic process to a sufficiently diverse stochastic process, yet with output that fits the class description.  
\item [(ii)] Top-$p$ sampling- In Top-$p$ sampling or nucleus sampling, the selection pool for the next token is determined by the cumulative probability of the most probable tokens. Setting a threshold $p$, the model includes just enough of the most probable tokens so that their combined probability reaches or exceeds this threshold. Again, the distribution mass is redistributed among these tokens and then the next token is sampled using this distribution. In Figure \ref{fig:digits} we see different examples of the digits `3' and `8' generated using the Top-$0.6$ method.  

\end{enumerate}

\section{Experimental results} \label{sec:experiment}

We conducted experiments on the MNIST and FashionMNIST datasets. Here are some details:
\begin{itemize}
    \item The images in both datasets were padded with zeros to $M\times M=32\times 32$, where $M=2^m$, $m=5$ and normalized to have values within $[0, 1]$. 
    \item We used the Haar wavelet basis for the MNIST images and the bior4.4 wavelet basis for the FashionMNIST. 
    \item The images were tokenized with a final threshold of $T=2^{-3}$ for MNIST and $T=2^{-4}$ for FashionMNIST.
    \item The maximal token sequence lengths were 1742 for MNIST and 3098 for FashionMNIST. 
    \item We trained two separate distillgpt2 models from scratch on the two datasets. As for the training configurations, both training sessions had batch size 4, learning rate 0.0004, and weight decay 0.01. 
    \item Models were trained on an NVIDIA A100 GPU with 80GB; MNIST occupied around 22GB while FashionMNIST occupied 61GB. Both models were trained for a few days.
    \end{itemize} 
    
    Results with different controlling methods appear below in Figures~\ref{fig:digits} and~\ref{fig:sandals}.

\begin{figure}[H]
\centering
\subfigure{\includegraphics[width=0.3\textwidth]{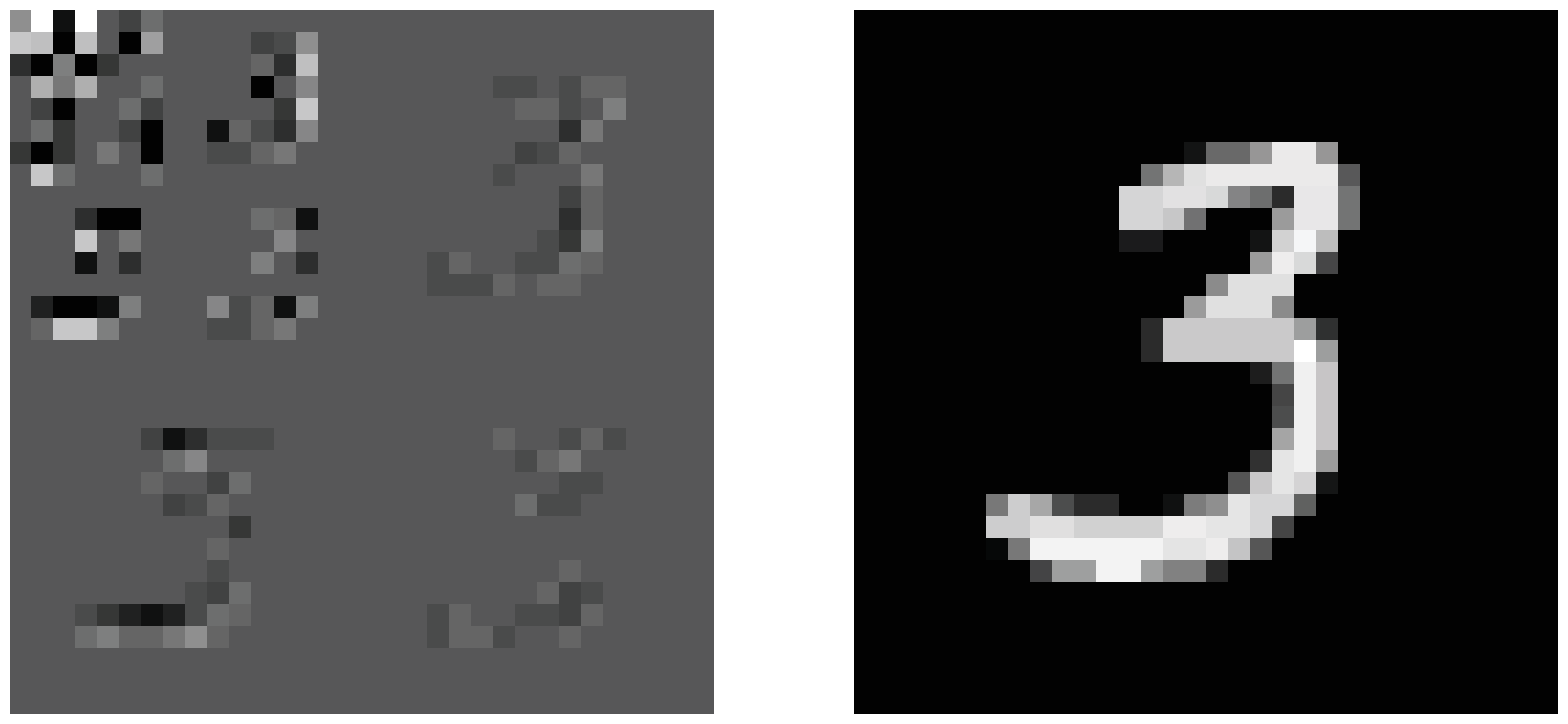}}
\hfill
\subfigure{\includegraphics[width=0.3\textwidth]{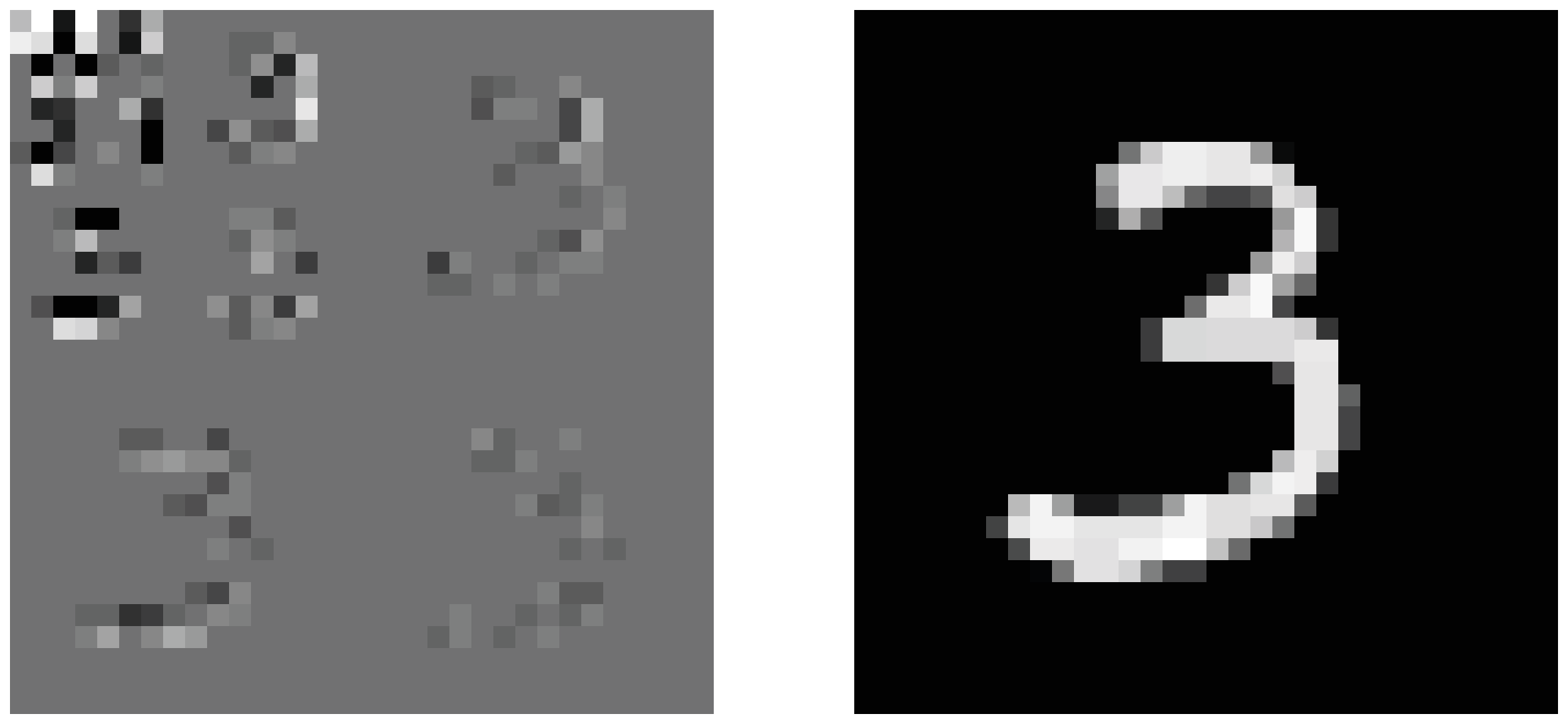}}
\hfill
\subfigure{\includegraphics[width=0.3\textwidth]{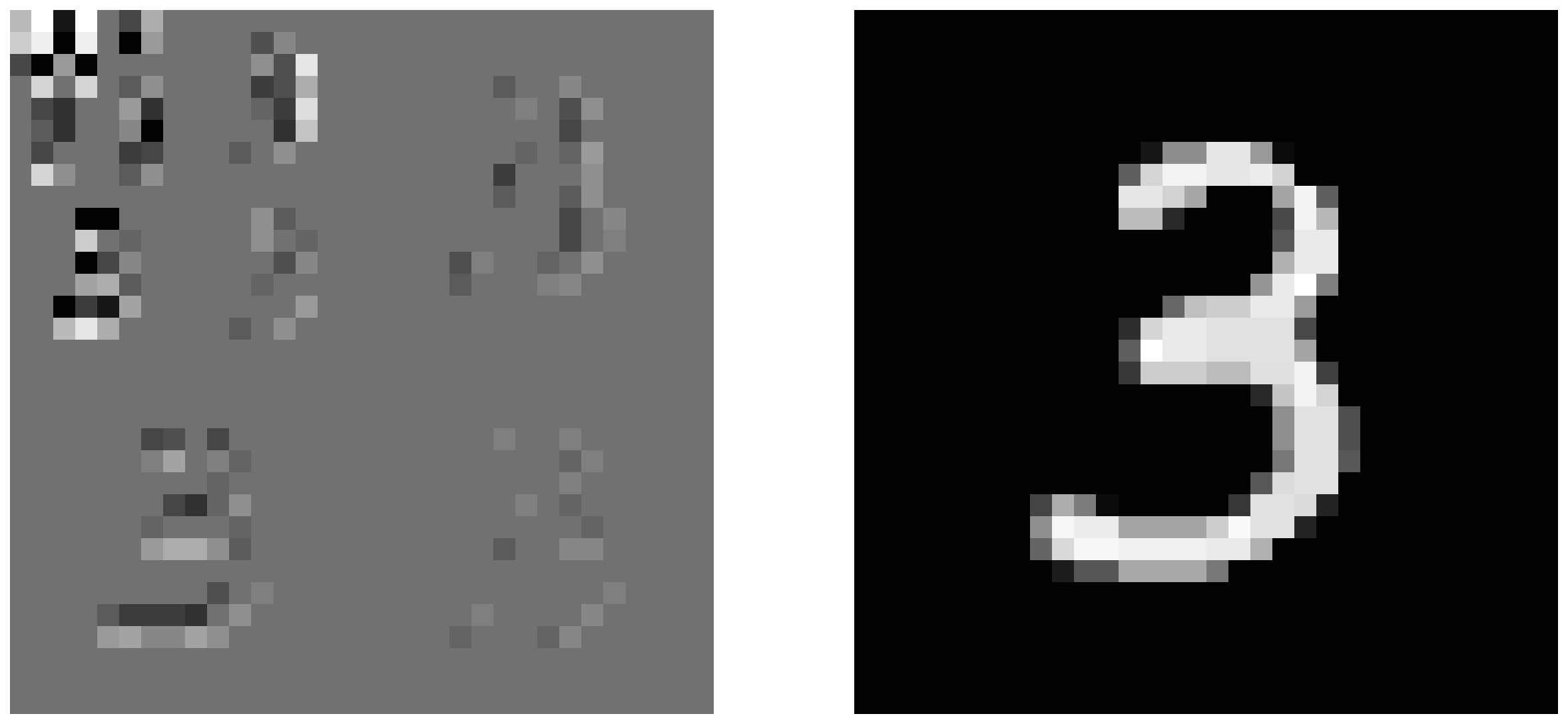}}
\vskip\baselineskip
\subfigure{\includegraphics[width=0.3\textwidth]{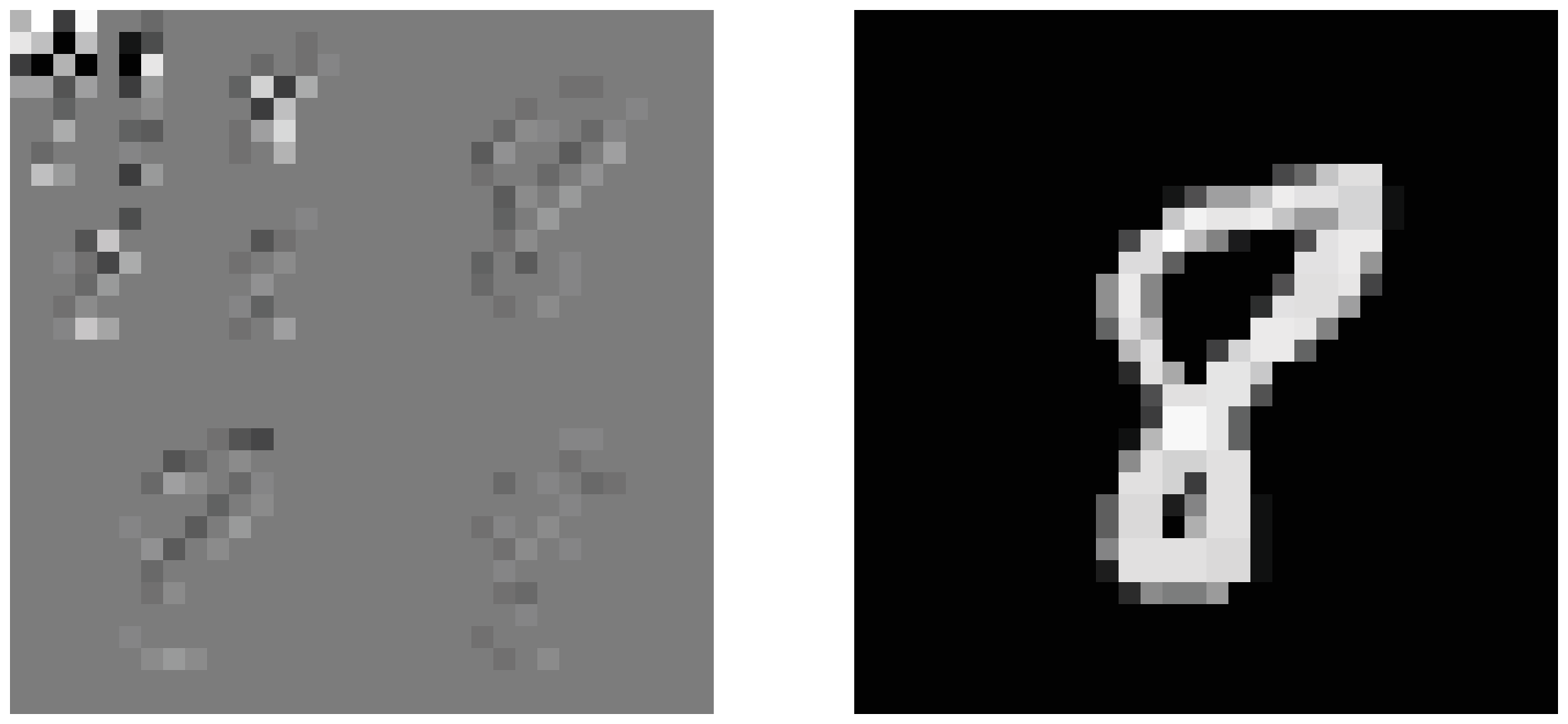}}
\hfill
\subfigure{\includegraphics[width=0.3\textwidth]{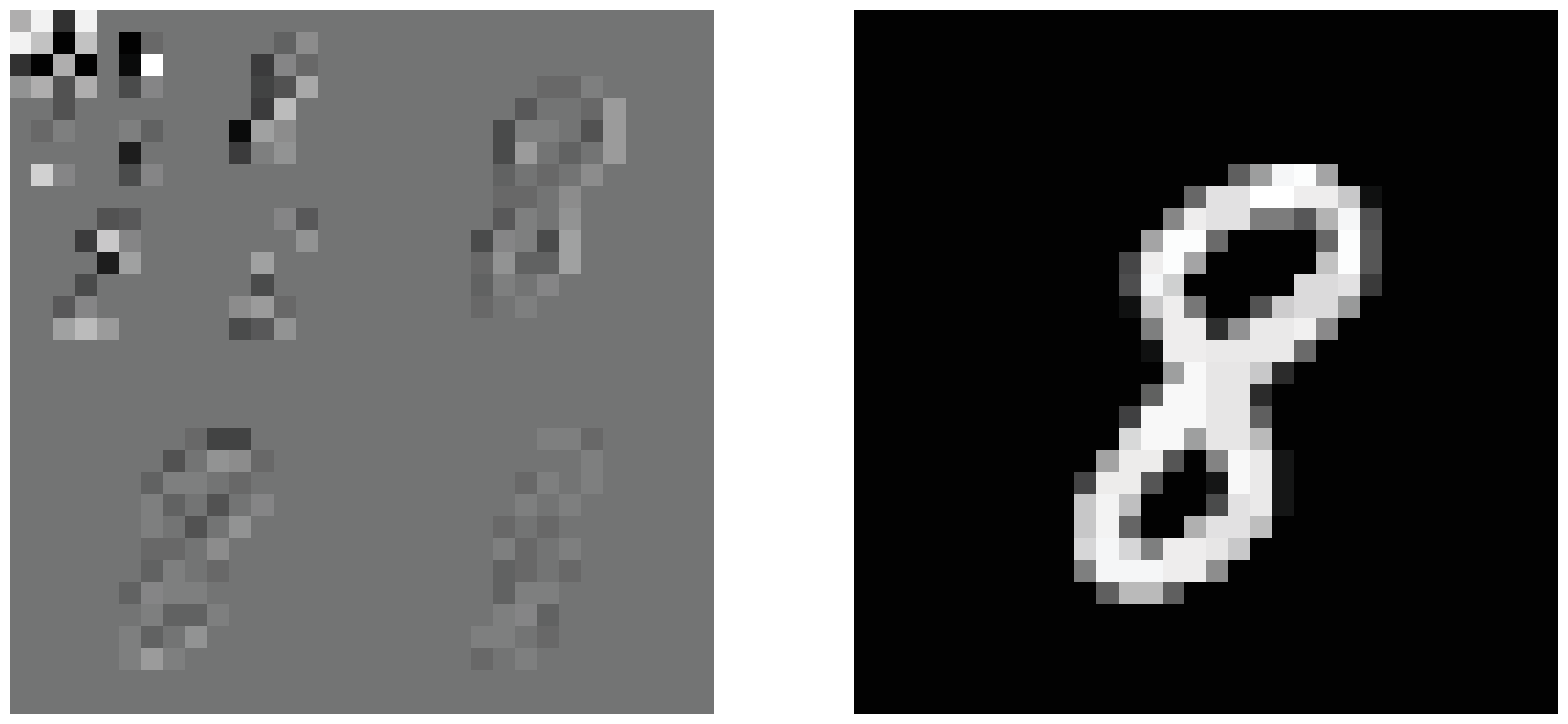}}
\hfill
\subfigure{\includegraphics[width=0.3\textwidth]{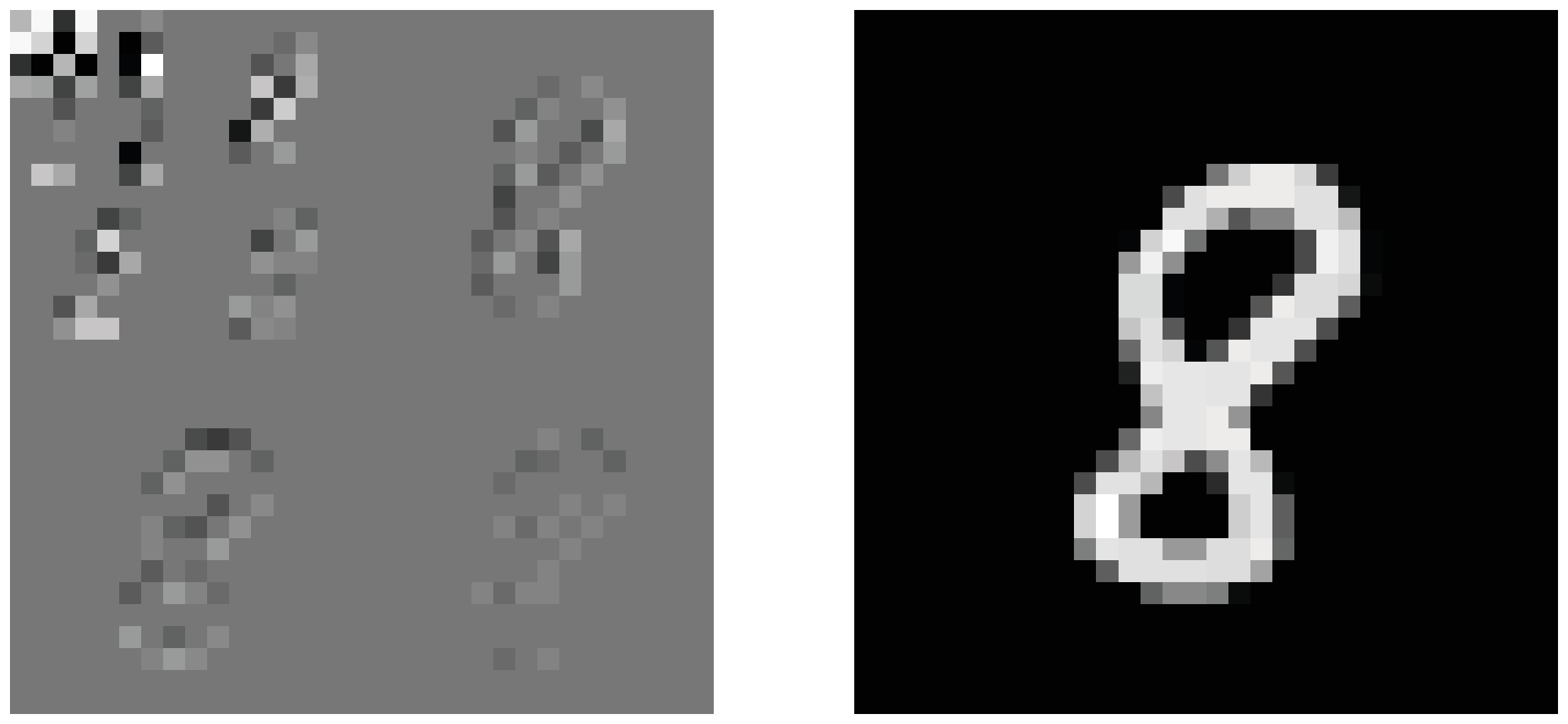}}
\caption{Digits generated with Top-$p=0.6$  along with a depiction of the generated wavelet coefficients.}
\label{fig:digits}
\end{figure}

\begin{figure}[H]
\centering
\subfigure{\includegraphics[width=0.3\textwidth]{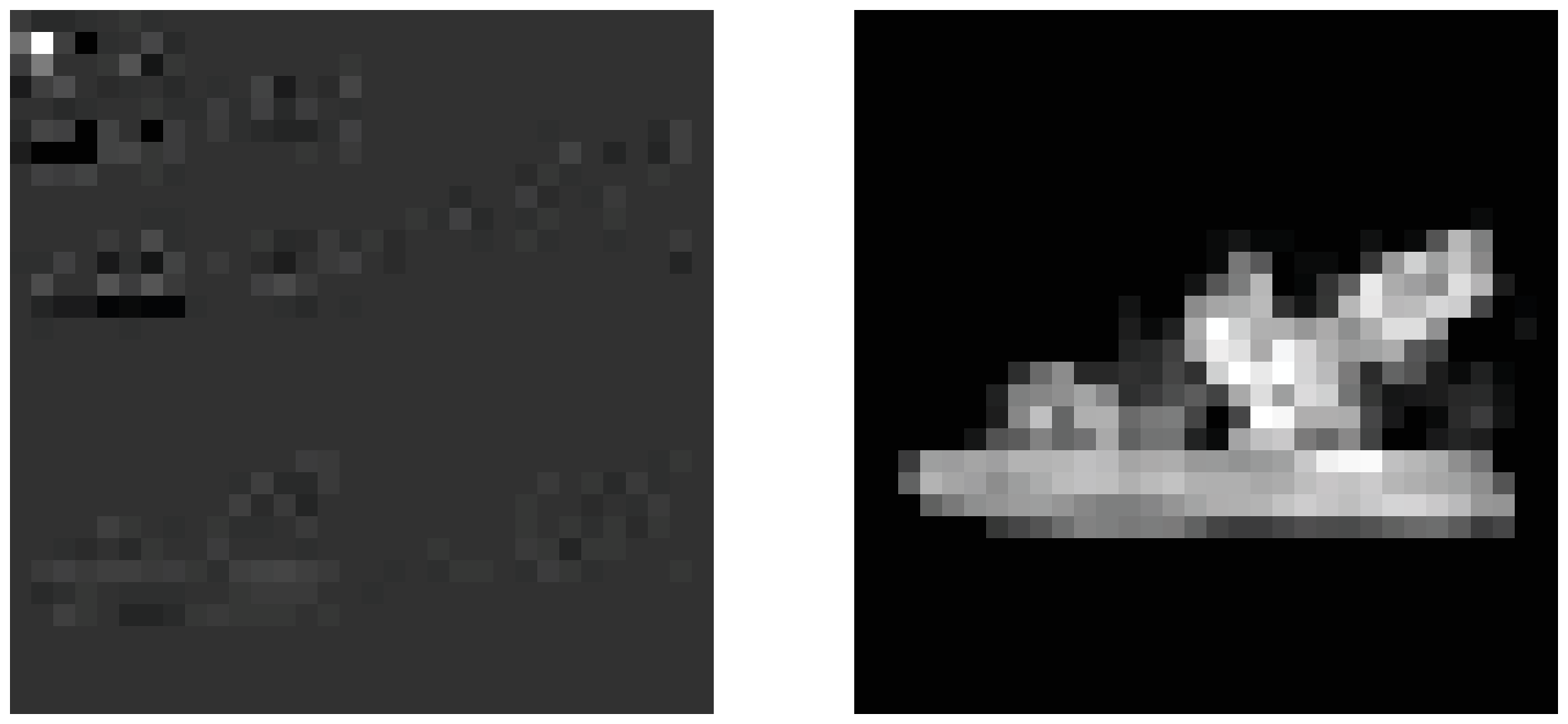}}
\hfill
\subfigure{\includegraphics[width=0.3\textwidth]{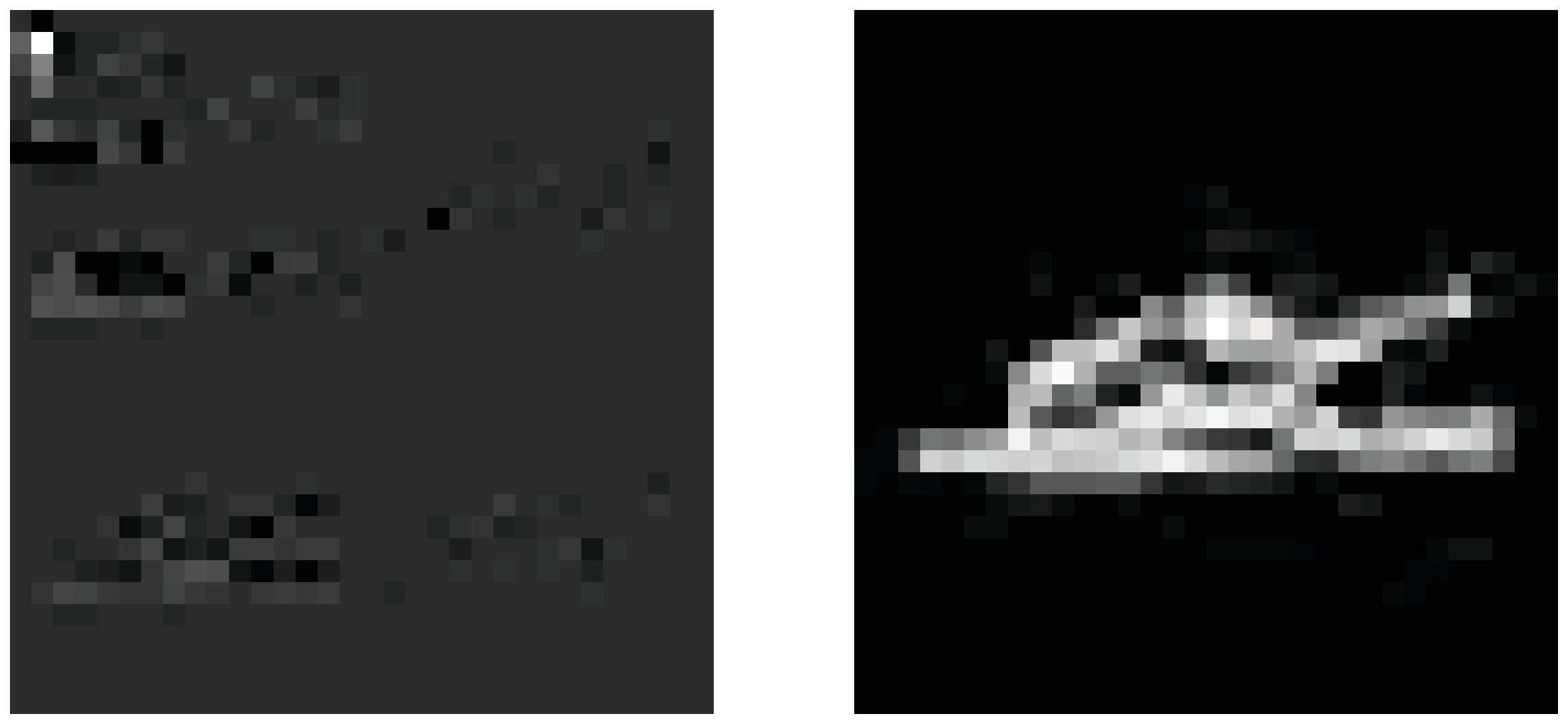}}
\hfill
\subfigure{\includegraphics[width=0.3\textwidth]{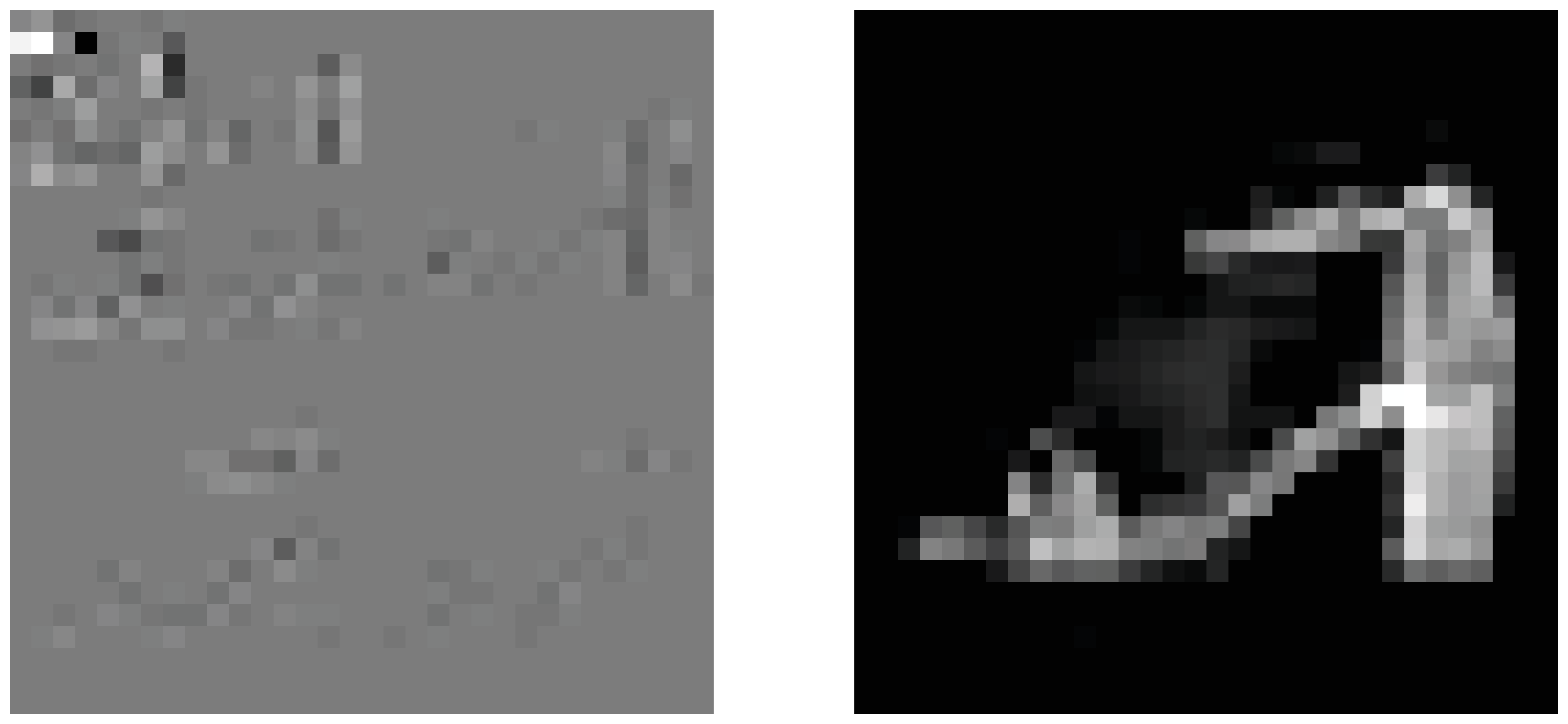}}
\vskip\baselineskip
\subfigure{\includegraphics[width=0.3\textwidth]{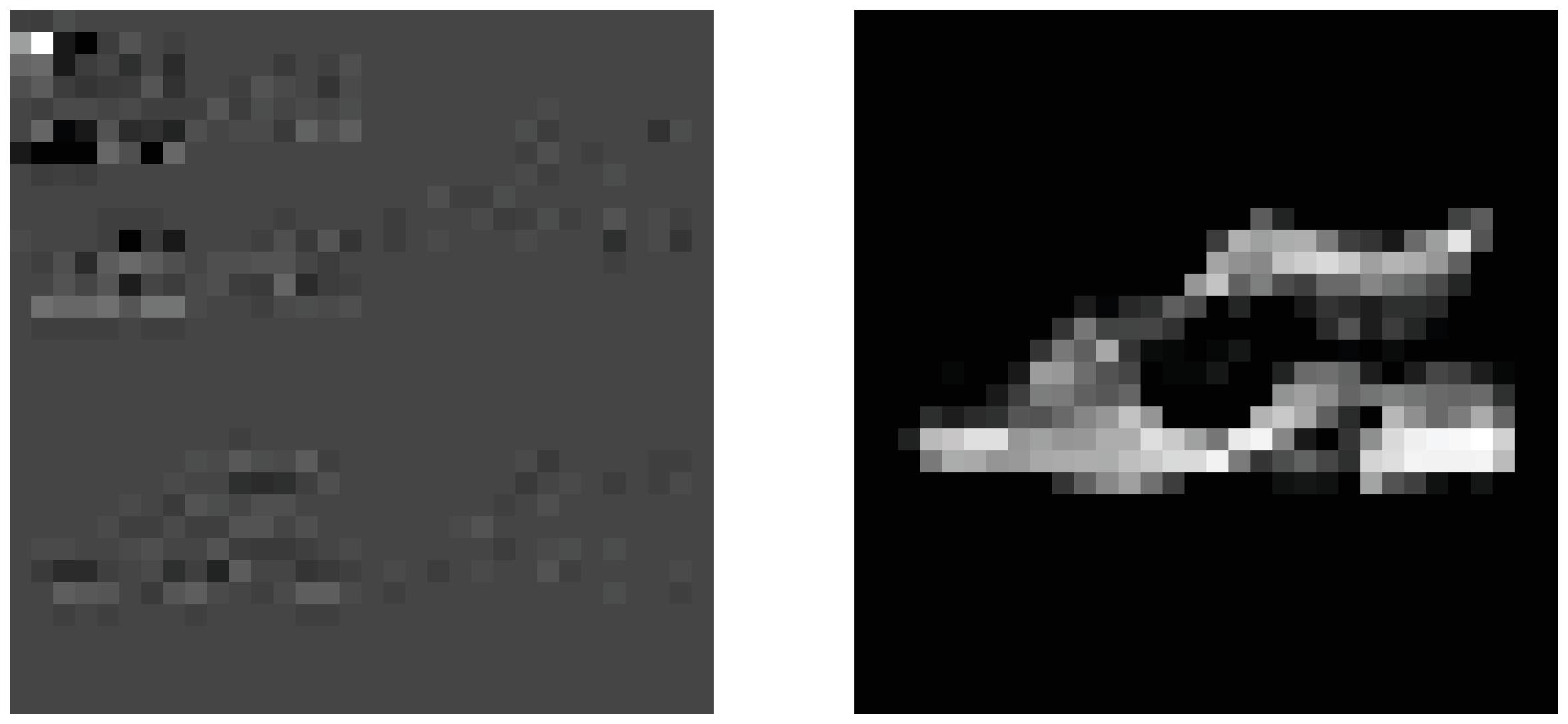}}
\hfill
\subfigure{\includegraphics[width=0.3\textwidth]{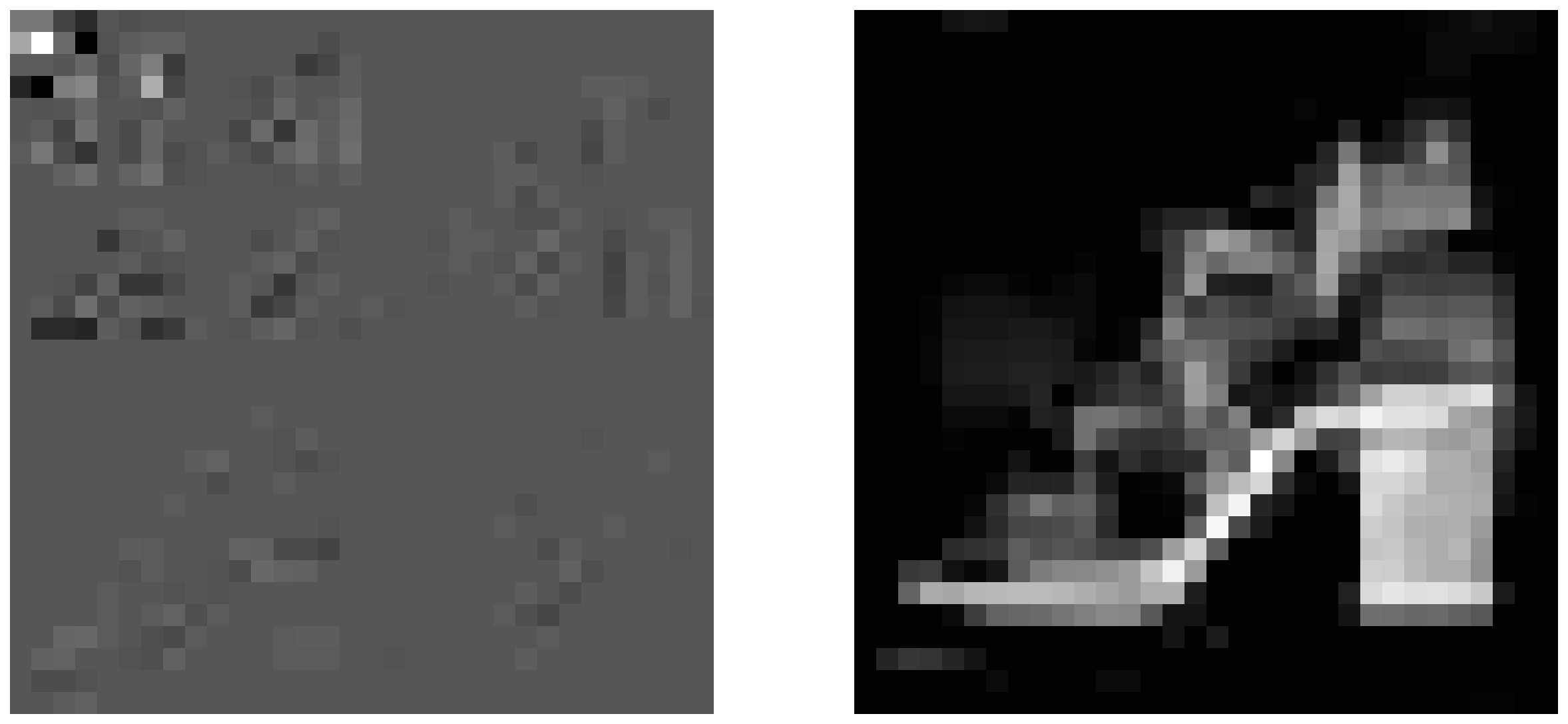}}
\hfill
\subfigure{\includegraphics[width=0.3\textwidth]{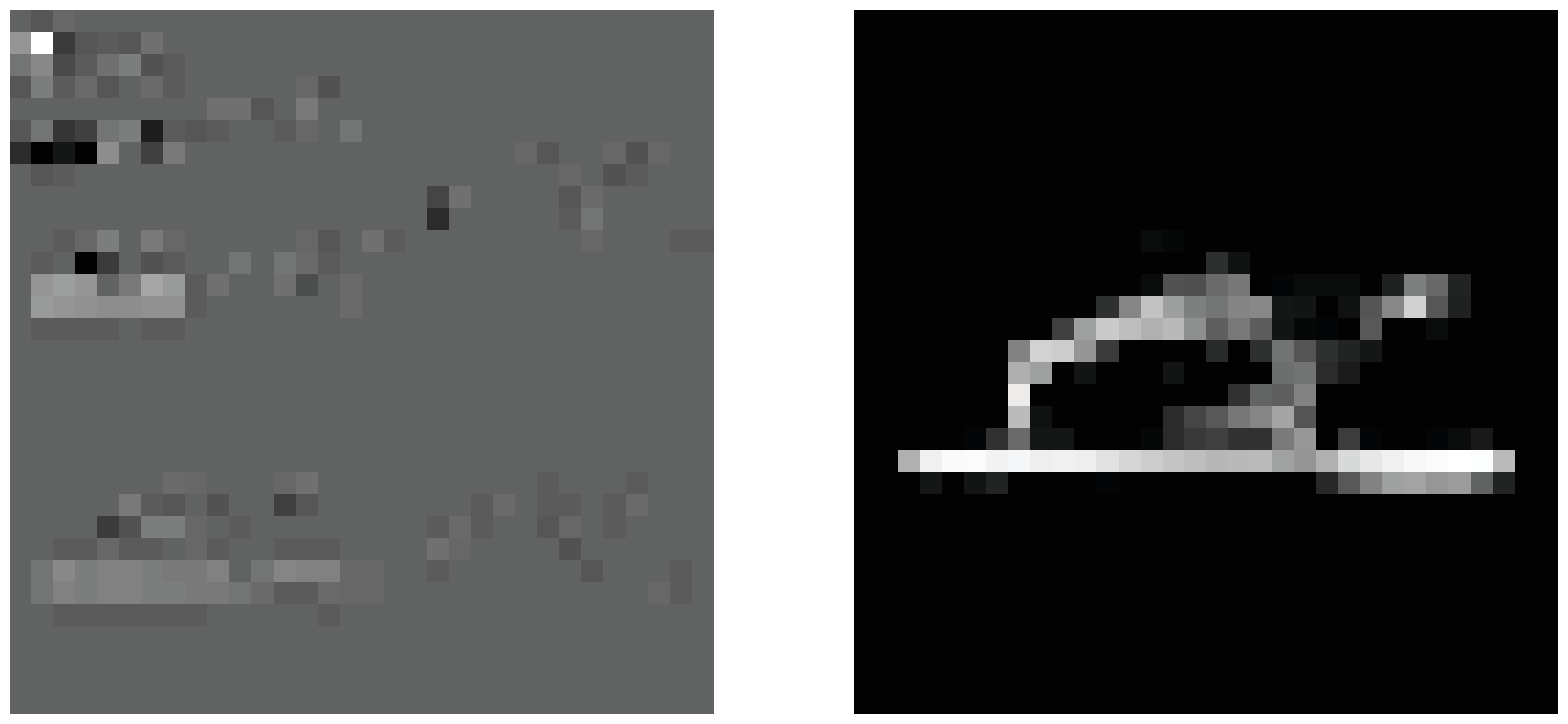}}
\caption{Sandals generated with Top-$k=2$ along with a depiction of the generated wavelet coefficients.}
\label{fig:sandals}
\end{figure}

More generated images for different classes of MNIST and FashionMNIST appear in the following figures.

\begin{figure}[H]
\centering
\subfigure{\includegraphics[width=0.3\textwidth]{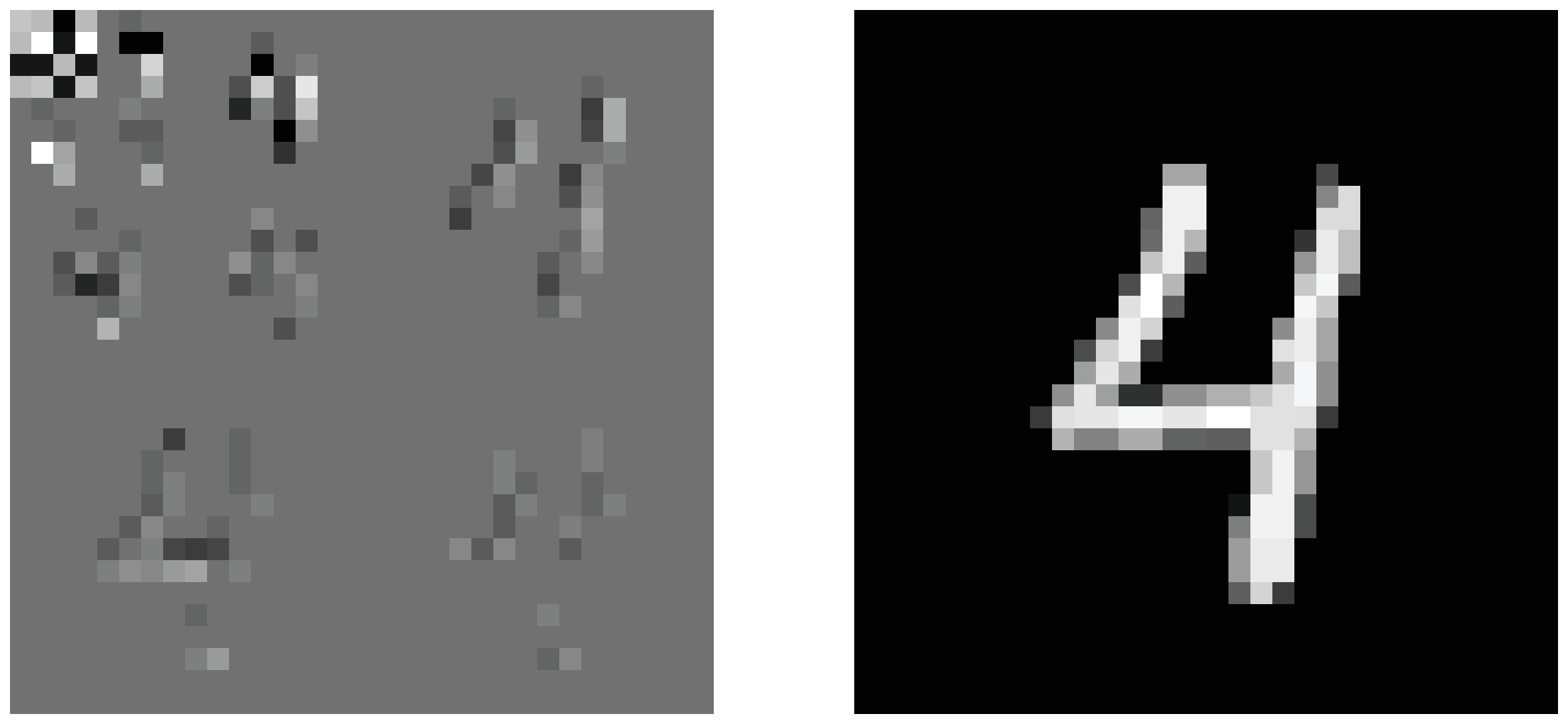}}
\hfill
\subfigure{\includegraphics[width=0.3\textwidth]{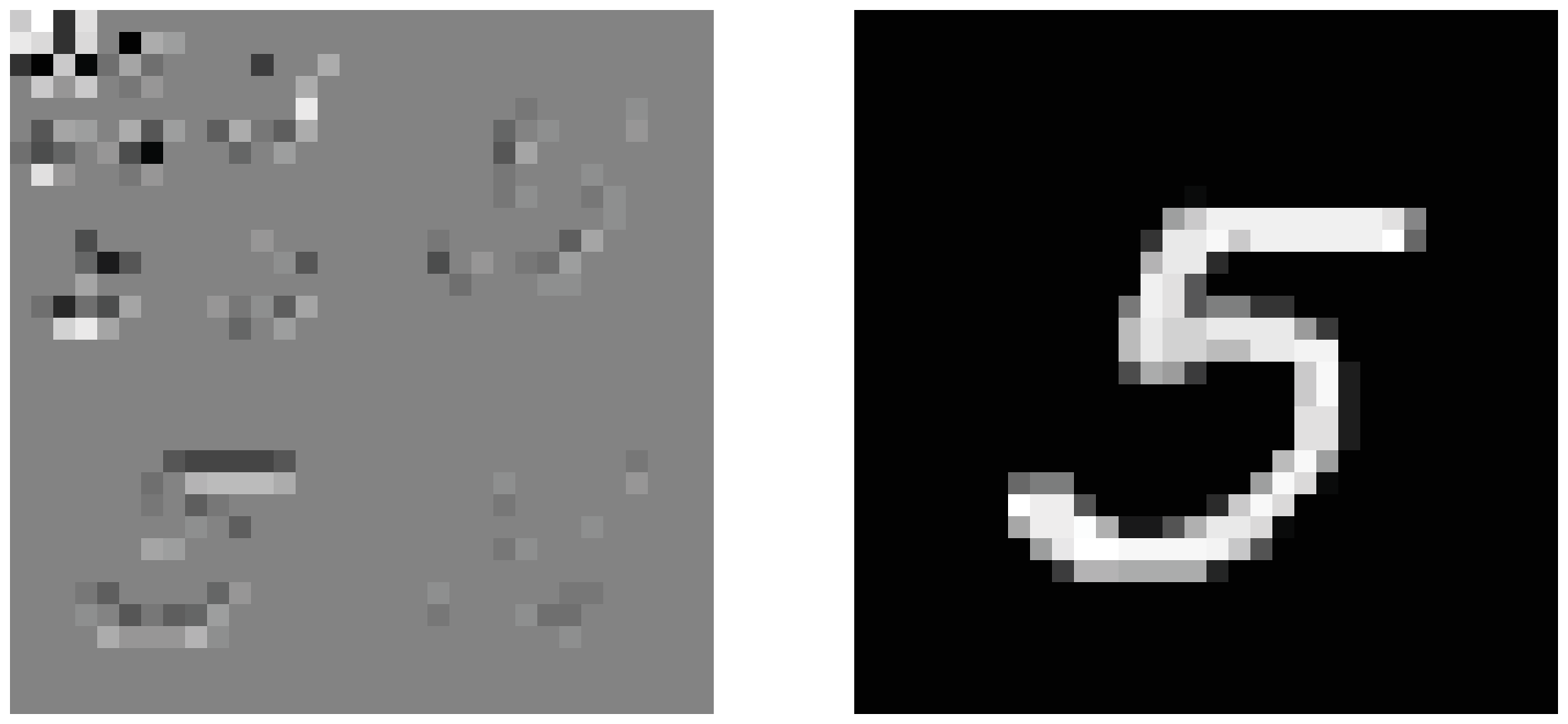}}
\hfill
\subfigure{\includegraphics[width=0.3\textwidth]{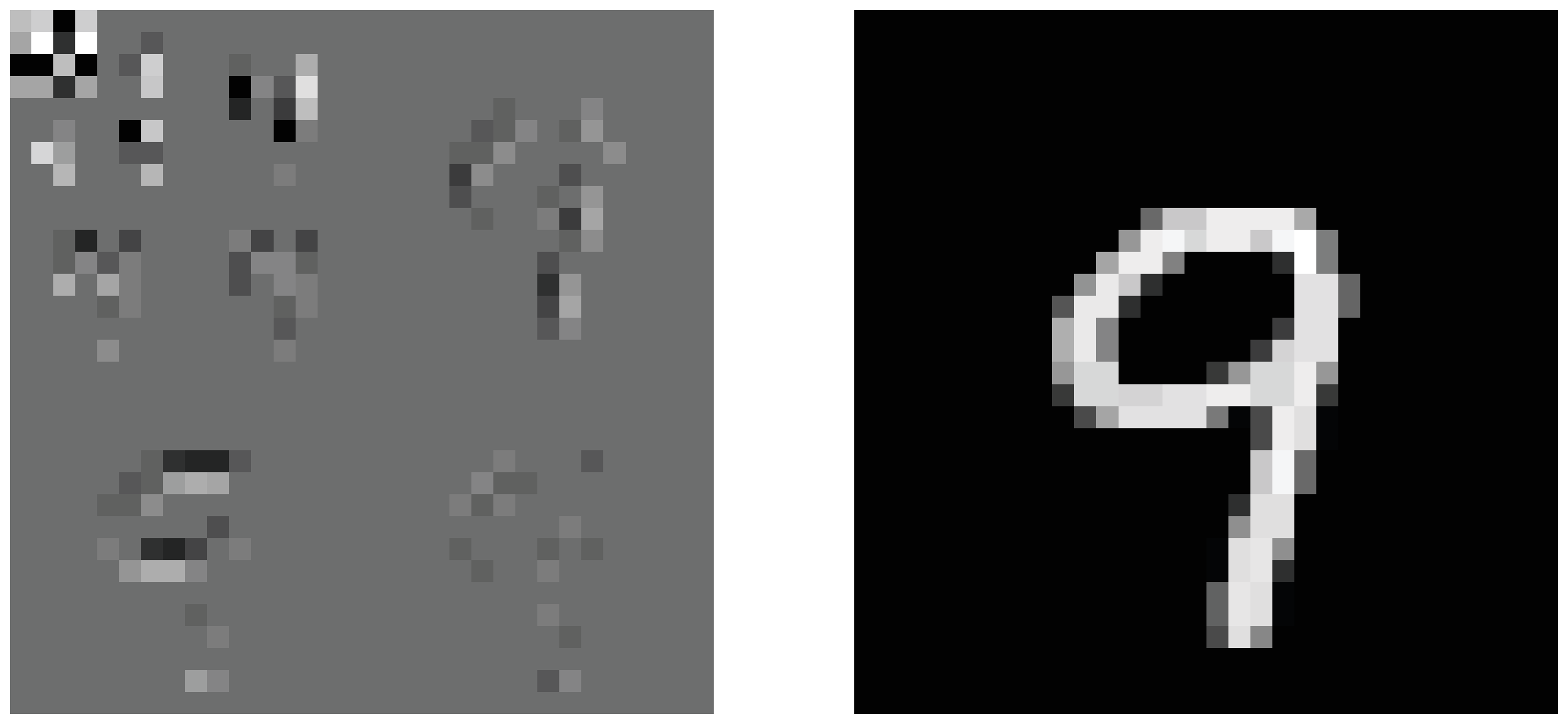}}
\caption{More MNIST results.}
\label{fig:moreMNIST}
\end{figure}

\begin{figure}[H]
\centering
\subfigure{\includegraphics[width=0.3\textwidth]{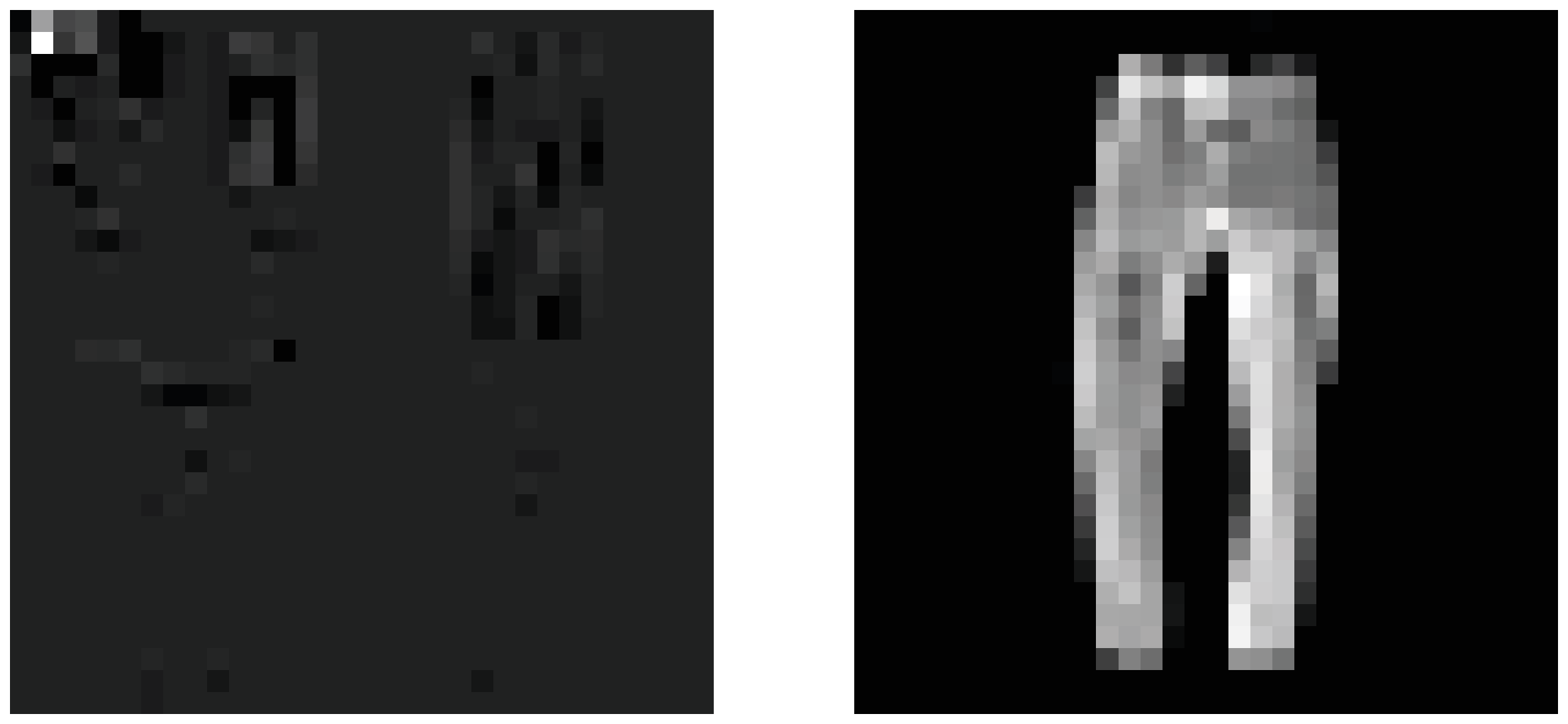}}
\hfill
\subfigure{\includegraphics[width=0.3\textwidth]{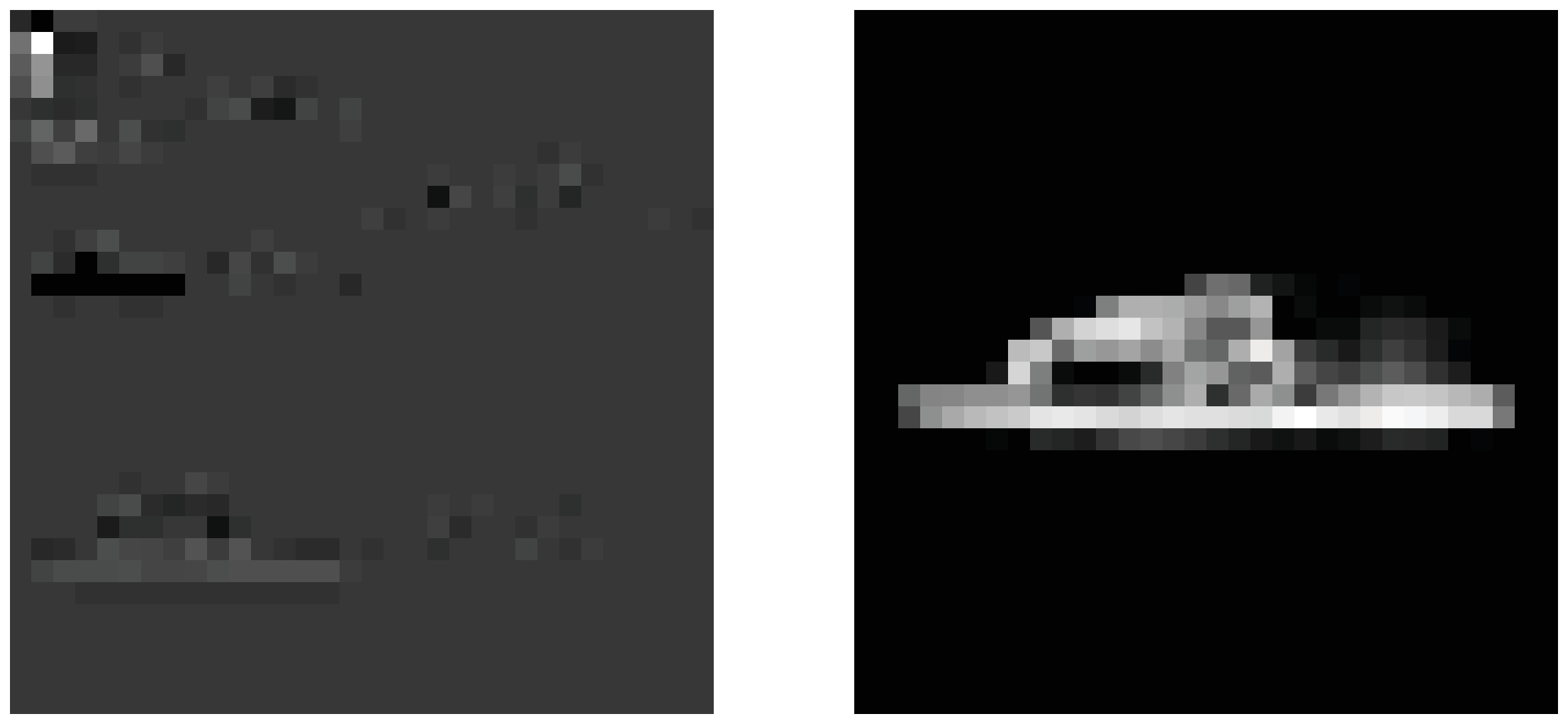}}
\hfill
\subfigure{\includegraphics[width=0.3\textwidth]{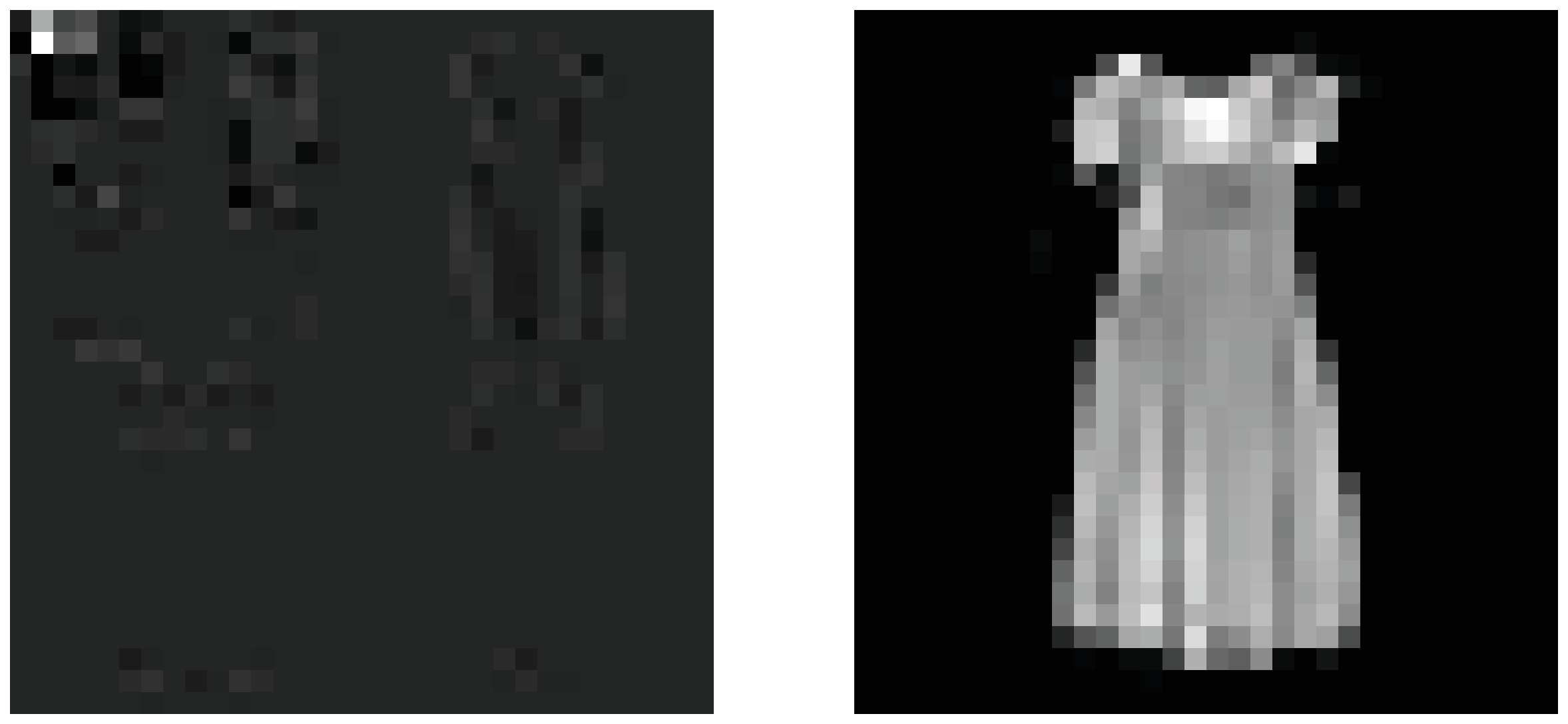}}
\caption{More FashionMNIST results.}
\label{fig:moreFashionMNIST}
\end{figure}

\section{Discussion and future work} \label{sec:discuss}

In this paper, we introduced a novel method for image generation that is based on elements of wavelet image coding and  NLP transformers. Unfortunately, our research group does not have access to sufficient computational resources at the moment, so this work serves as a first modest proof of concept. Indeed, the wavelet representation is a powerful tool in image processing that can serve as a basis for many image generation functionalities. Here, we list some directions that we will consider for future work.

\subsection{Generation of color images at high resolution and with fine details}

In our experiments, we only generated small grayscale images. We provide here some details on how the method can be generalized:
\begin{enumerate}
    \item [(i)] Color images - For color images (or even spectral images), we may adopt a well-known paradigm from image compression. For improved performance, one may transform input images in the $RGB$ color space to the $YCbCr$ color space. The $Y$ component is the luminance component, essentially the image's grayscale part. The other two components, $Cb$ and $Cr$, capture the color information of the image. Typically, the luminance component carries most of the visual information, and thus also, its encoding is usually the significant part of an encoded image. In image coding, one usually encodes separately each of the three channels. Our method can then be generalized to color images by applying the DWT and the tokenization process separately to each color channel. 
    \item[(ii)] Generating fine details - Using our wavelet model, finer details are captured at higher bit-planes. The choice of the final threshold of the final bit-plane provides excellent and very consistent control over the amount of detail one wishes to generate. This quantization technique is at the heart of the JPEG algorithms and translates to very specific modes in digital cameras that can be set to: ``Visually Lossless'', ``High'', ``Medium'', etc. This exact form of control also applies to wavelets but, unfortunately, is not the default mode of operation in JPEG2000. Obviously, to generate finer details, one needs to train the transformer on longer token sequences, again requiring more computational resources.   
    \end{enumerate}

\subsection{Support for generation of compositions of blobs} \label{subsec:blob}

In many cases, one wishes to apply fine-grained control of compositional text-to-image generation, where certain locations in the image, marked perhaps with bounding boxes or ellipses, receive different textual descriptions \cite{nie2024compositional}. One possible method to accomplish this using the wavelet generative approach is to apply the transformer in evaluation mode and apply the vector representation of the blob's textual prompt as described in Subsection \ref{subsec:guide} whenever the bit-plane scan is at indices of wavelet coefficients whose support intersects the blob. 

\subsection{Multi-modal generation}

The ability to represent an image's visual information as a sequence of tokens presents an attractive possibility of merging the wavelet-based tokens with other language tokens to create a unified multi-modal transformer.

\section*{Funding}
N. Sharon is partially supported by the NSF-BSF award 2019752. W. Mattar is partially supported by The Nehemia Levtzion Scholarship for Outstanding Doctoral Students from the Periphery (2023). N. Sharon and W. Mattar are partially supported by the DFG award
514588180.

\bibliographystyle{unsrtnat}

\begin{thebibliography}{99}

\bibitem{buccigrossi1999image}
R. W. Buccigrossi and E. P. Simoncelli,
{\it Image compression via joint statistical characterization in the wavelet domain},
IEEE transactions on Image processing {\bf 8} (1999), 1688-1071. 


\bibitem{FLAN-T5} 
H. W. Chung, L. Hou, S. Longpre, B. Zoph, Y. Tay, W. Fedus, Y. Li, X. Wang, M. Dehghani, S. Brahma, A. Webson, S. S. Gu, Z. Dai, M. Suzgun, X. Chen, A. Chowdhery, A. Castro-Ros, M. Pellat, K. Robinson, D. Valter, S. Narang, G. Mishra, A. Yu, V. Zhao, Y. Huang, A. Dai, H. Yu, S. Petrov, E. H. Chi, J. Dean, J. Devlin, A. Roberts, D. Zhou, Q. V. Le and J. Wei, {\it Scaling Instruction-Finetuned Language Models}, Journal of Machine Learning Research {\bf 25} (2024), 1-153. 

\bibitem{Cern2011DISCRETEC9}
C. Dana and F. V{\'a}clav,
{\it DISCRETE {CDF} 9 / 7 WAVELET TRANSFORM FOR FINITE-LENGTH SIGNALS},
https://api.semanticscholar.org/CorpusID:208013335, 2011.

\bibitem{Daubechies}
I. Daubechies, {\it Ten Lectures on Wavelets}, SIAM, 1992.

\bibitem{deng2012mnist}
L. Deng, {\it The mnist database of handwritten digit images for machine learning research},
IEEE signal processing magazine {\bf 29} (2012), 141-142. 

\bibitem{DeVore}
R. DeVore, {\it Nonlinear approximation}, Acta Numerica {\bf 7} (1998), 51-150.

\bibitem{dhariwal2021diffusion}
P. Dhariwal and A. Nichol,
{\it Diffusion models beat gans on image synthesis},
Advances in neural information processing systems {\bf 34} 2021, 8780-8794.

\bibitem{esser2021taming}
P. Esser, R. Rombach and B. Ommer,
{\it Taming transformers for high-resolution image synthesis},
Proceedings of the IEEE/CVF conference on computer vision and pattern recognition 2021,
12873-12883.

\bibitem{fan-etal-2018-hierarchical}
A. Fan, M. Lewis and Y. Dauphin, 
{\it Hierarchical Neural Story Generation},
Proceedings of the 56th Annual Meeting of the Association for Computational Linguistics (Volume 1: Long Papers) 2018, 889-898.

\bibitem{BPE}
P. Gage, {\it A new algorithm for data compression}
The C Users journal archive {\bf 12} (1994), 23-38.


\bibitem{guth2022wavelet}
F. Guth, S. Coste, V. De Bortoli and S. Mallat,
{\it Wavelet score-based generative modeling},
 Advances in Neural Information Processing Systems {\bf 35} (2022),
478-491.

\bibitem{NEURIPS2020_4c5bcfec}
J. Ho, A. Jain and P. Abbeel,
{\it Denoising Diffusion Probabilistic Models}, 
Advances in Neural Information Processing Systems {\bf 33} (2020), 6840-6851.

\bibitem{distilGPT2}
{\it Hugging{F}ace {DistilGPT2}},
https://huggingface.co/distilbert/distilgpt2, 2019.

\bibitem{hugtoken}
{\it Hugging{F}ace Tokenizer}
https://huggingface.co/docs/tokenizers/index.

\bibitem{Jiang_2021_ICCV}
L. Jiang, B. Dai, W. Wu and C. C. Loy,
{\it Focal Frequency Loss for Image Reconstruction and Synthesis},
Proceedings of the IEEE/CVF International Conference on Computer Vision (ICCV) 2021, 13919-13929.

\bibitem{803428}
M. M. Kivanc, I. Kozintsev, K. Ramchandran and P. Moulin,
{\it Low-complexity image denoising based on statistical modeling of wavelet coefficients}, 
IEEE Signal Processing Letters {\bf 6} (1999), 300-303.

\bibitem{Mallat}
S. Mallat, {\it A Wavelet tour of signal processing, the sparse way},
Academic Press, 2009.

\bibitem{mihcak1999spatially}
K. M. Mihcak, I. Kozintsev and K. Ramchandran,
{\it Spatially adaptive statistical modeling of wavelet image coefficients and its application to denoising}, proceedings of 1999 IEEE International Conference on Acoustics, Speech, and Signal Processing ICASSP99  {\bf 6} (1999), 3253-3256.

\bibitem{pmlr-v162-nichol22a}
A. Q. Nichol, P. Dhariwal, A. Ramesh, P. Shyam, P. Mishkin, B. Mcgrew, I. Sutskever and M. Chen, 
{\it {GLIDE}: Towards Photorealistic Image Generation and Editing with Text-Guided Diffusion Models},
Proceedings of the 39th International Conference on Machine Learning {\bf 162} (2022), 16784-16804. 

\bibitem{waveletdiffusion1}
H. Phung, Q. Dao and A. Tran,  
{\it Wavelet Diffusion Models are fast and scalable Image Generators},
IEEE/CVF Conference on Computer Vision and Pattern Recognition (CVPR) 2023, 10199-10208.

\bibitem{FormalTrans}
M. Phuong and M. Hutter, {\it Formal Algorithms for Transformers}, 
  https://arxiv.org/abs/2207.09238, 
 2022.


\bibitem{radford2021learning}
A. Radford, J. W.  Kim, C. Hallacy, A. Ramesh,G. Goh, S. Agarwal, G. Sastry, A. Askell, P. Mishkin, J. Clark and others,
{\it Learning transferable visual models from natural language supervision},
International conference on machine learning (2021), 8748-8763.


\bibitem{ramesh2022hierarchical}
A. Ramesh, P. Dhariwal, A. Nichol, C. Chu andM. Chen,
{\it Hierarchical text-conditional image generation with clip latents},
arXiv, 2022.

\bibitem{ramesh2021zero}
A. Ramesh,M. Pavlov, G. Goh, S. Gray, C. Voss, A. Radford, M.  Chen and I. Sutskever,
{\it Zero-shot text-to-image generation},
International conference on machine learning 2021, 8821-8831.


\bibitem{rombach2022high}
R. Rombach, A. Blattmann D. Lorenz, P. Esser and B. Ommer,
{\it High-resolution image synthesis with latent diffusion models},
Proceedings of the IEEE/CVF conference on computer vision and pattern recognition 2022,
10684-10695.

\bibitem{SPIHT}
A. Said and W. Pearlman,
{\it A new, fast, and efficient image codec based on set partitioning in hierarchical trees},
IEEE Transactions on Circuits and Systems for Video Technology {\bf 6} (1996), 243-250.

\bibitem{saharia2022photorealistic}
C. Saharia, W. Chan, S. Saxena, L. Li, J. Whang, E. Denton, K. Ghasemipour, L. Gontijo, Raphael, A. Karagol, Burcu, Salimans, Tim and others,
{\it Photorealistic text-to-image diffusion models with deep language understanding},
Advances in neural information processing systems {\bf 35} (2002), 
36479-36494.

\bibitem{sanh2019distilbert}
V. Sanh, {\it DistilBERT, a distilled version of BERT: smaller, faster, cheaper and lighter},
Proceedings of Thirty-third Conference on Neural Information Processing Systems (NIPS2019).


\bibitem{EZW}
J. Shapiro, {\it Embedded image coding using zerotrees of wavelet coefficients},
IEEE Transactions in signal processing {\bf 41} 1993, 3445-3462.


\bibitem{JPEG2000Book}
D. Taubman and M. Marcellin, {\it JPEG2000: Image Compression Fundamentals, Standards and Practice, 2nd edition}, Springer, 2002.


\bibitem{tay2022efficient}
Y. Tay, M. Dehghani, D. Bahri and D. Metzler,
 {\it Efficient transformers: A survey},
 ACM Computing Surveys {\bf 55} (2022), 1-28.


\bibitem{tian2024visual}
K. Tian, Y.  Jiang, Z. Yuan, B. Peng, and L. Wang,
{\it Visual Autoregressive Modeling: Scalable Image Generation via Next-Scale Prediction},
arXiv 2024. 



\bibitem{vaswani2017attention}
A. Vaswani, N. Shazeer, N. Parmar, J. Uszkoreit, L. Jones, A. Gomez, L. Kaiser and I. Polosukhin, {\it Attention is all you need},
Advances in neural information processing systems {\bf 30} (2017).

\bibitem{wang2023images}
X. Wang,W. Wang, Y. Cao, C. Shen  and T. Huang, 
{\it Images speak in images: A generalist painter for in-context visual learning},
Proceedings of the IEEE/CVF Conference on Computer Vision and Pattern Recognition 2023,
6830-6839.

\bibitem{lee2022autoregressive}
D. Lee, C. Kim, S. Kim, M. Cho and W. Wook,
{\it Autoregressive image generation using residual quantization},
 Proceedings of the IEEE/CVF Conference on Computer Vision and Pattern Recognition 2022,
11523-11532.

\bibitem{zhu2023wdig}
 Q. Zhu,X. Li, J. Sun and H. Bai,
{\it WDIG: a wavelet domain image generation framework based on frequency domain optimization},
EURASIP Journal on Advances in Signal Processing 2023, 66.




\bibitem{Wallace}
G. K. Wallace, {\it The JPEG still picture compression standard},
IEEE Transactions on Consumer Electronics {\bf 38} (1992), xviii-xxxiv.

\bibitem{nie2024compositional}
N. Weili, L. Sifei, M. Morteza, L. Chao, E. Benjamin and V. Arash, 
{\it Compositional Text-to-Image Generation with Dense Blob Representations}, 
arXiv 2024. 

\bibitem{32tokens}
Q. Yu, M. Weber, X. Deng, X. Shen, D. Cremers and L. Chen,
{\it An image is worth 32 tokens for reconstruction and generation},
 NeurIPS 2024, to appear.

\bibitem{yu2021wavefill}
Y. Yu, F. Zhan, S. Lu, J. Pan, F. Ma, X. Xie and C. Miao, Chunyan,
{\it Wavefill: A wavelet-based generation network for image inpainting},
Proceedings of the IEEE/CVF international conference on computer vision 2021,
14114-14123.

\end{thebibliography}

\end{document}